\documentclass[sigconf, nonacm]{acmart}

\newcommand\vldbpagestyle{plain} 

\usepackage{tcolorbox}
\usepackage{enumitem}
\usepackage{listings}
\lstnewenvironment{sqllist}[1][]
{\lstset{
    language=SQL,
    basicstyle=\ttfamily,
    breaklines=true,
    gobble=2,
    #1
  }}
{}

\lstnewenvironment{prompt}[1][]
{\lstset{
    basicstyle=\ttfamily,
    numbers=none,
    keywords={},
    #1
  }}
{}

\newcommand{\sql}[1]{\texttt{"{#1}"}}

\usepackage{longtable}
\usepackage{booktabs}
\usepackage{multirow}

\usepackage{bbm}
\usepackage{amsmath}
\DeclareMathOperator*{\argmax}{arg\,max}

\theoremstyle{definition}

\theoremstyle{definition}

\usepackage{xcolor}

\begin{document}
\title{Fundamental Challenges in Evaluating Text2SQL Solutions and Detecting Their Limitations}

\author{Cedric Renggli$^{1}$, Ihab F. Ilyas$^{2*}$, Theodoros Rekatsinas$^{1}$}
\thanks{*Work done while at Apple}
\affiliation{%
\institution{$^{1}$Apple,~~$^{2}$University of Waterloo}
}

\begin{abstract}

In this work, we dive into the fundamental challenges of evaluating Text2SQL solutions and highlight potential failure causes and the potential risks of relying on aggregate metrics in existing benchmarks. We identify two largely unaddressed limitations in current open benchmarks: (1) data quality issues in the evaluation data, mainly attributed to the lack of capturing the probabilistic nature of translating a natural language description into a structured query (e.g., NL ambiguity), and (2) the bias introduced by using different match functions as approximations for SQL equivalence.

To put both limitations into context, we propose a unified taxonomy of all Text2SQL limitations that can lead to both prediction and evaluation errors. We then motivate the taxonomy by providing a survey of Text2SQL limitations using state-of-the-art Text2SQL solutions and benchmarks. We describe the causes of limitations with real-world examples and propose potential mitigation solutions for each category in the taxonomy. We conclude by highlighting the open challenges encountered when deploying such mitigation strategies or attempting to automatically apply the taxonomy. 

\end{abstract}

\maketitle

\pagestyle{\vldbpagestyle}

\section{Introduction}

Text2SQL (or NL2SQL) is defined as the task of translating text in the form of a natural language (NL) description or question, along with the serialization of the corresponding schema of a database (DB), into a structured query language (SQL) query. Text2SQL models have significantly matured and shown increasing success in public benchmarks over the last two years, thanks to the improved NL understanding capabilities provided by modern large language models (LLMs) and the availability of larger training and evaluation data in public benchmarks. Nonetheless, to the best of our knowledge, there are few known successful applications of Text2SQL solutions in practice~\citep{koutrika2024natural, floratou2024nl2sql}. Most related work so far has focused on very specific limitations in existing trained Text2SQL models, such as their inability to provide accurate predictions, especially for complex schemata or queries~\citep{zhang2024benchmarking, gkini2021depth, mitsopoulou2024analysis}.
In this work, we ask ourselves two main questions: \textit{Q1: Are there other reasons, beyond model limitations, in the end-to-end Text2SQL solution that can result in inaccurate predictions?} And, \textit{Q2: Are there limitations in the evaluation procedure that can bias the accuracy estimation by over- or under-estimating Text2SQL solutions' capabilities?}
To answer both of these questions, we take a bird's-eye view of all limitations hampering high-accuracy Text2SQL solutions (i.e., the end-to-end system, including input preparation and result extraction and verification steps), beyond model limitations alone.  Furthermore, we realize that there are other fundamentally unaddressed challenges in the evaluation process used in existing benchmarks that can prevent them from providing reliable performance metrics.

We identify two such challenges: (a) existing benchmarks suffer from a significant fraction of data quality issues (or noise) concerning input features or ground truth labels, and (b) the match functions used to approximate SQL equivalence and compute aggregate performance metrics introduce biases into the measurement.

\paragraph{Data Quality Issues}
We observe two complementary issues reducing the data quality of existing evaluation artifacts (i.e., making them noisy): First, existing benchmark data are incomplete in that they ignore the probabilistic nature of the task by providing exactly one single ground truth label per input and thus ignoring alternative, equally likely SQL queries for the input. Second, some fraction of the data is inaccurate in that the provided ground truth label does not correspond to (any of) the most probable SQL queries. There are multiple reasons why there can be more than one valid SQL interpretation for an input tuple consisting of a natural language description and a serialized version of the schema. The two most important ones we identify are (a) inherent ambiguity in natural language, and (b) additional ambiguity due to information loss during the schema definition or serialization process. While the first aspect has been acknowledged in the NLP literature~\citep{jusoh2018study}, it, together with the second aspect, has been widely ignored in Text2SQL-related work and benchmarks~\cite{floratou2024nl2sql}. To understand (b), note that a \textit{relational schema}, as formally introduced in the database theory literature, contains all the information the schema entails, e.g., including \textit{integrity constraints}. To see the impact of losing some information in the definition or serialization process, consider a categorical attribute representing the status of a person (cf. Figure~\ref{fig:example_schema}) and taking values in `T' (true), `F' (false), or `S' (suspended). Such an attribute would typically be defined as a \verb|CHAR(1)| datatype in common SQL databases like SQLite. This definition of such an attribute, without specifying the exact domain, introduces additional ambiguity for any query asking about it, given that not even a human expert would know how these values are stored and have to be queried in the database without having access to the data.

\paragraph{Approximate SQL Equivalence}
SQL equivalence is well-defined in the database literature. Formally, two SQL queries are said to be equivalent for a fixed schema if there does not exist any valid database instance for which the two queries yield different results. It is well-known that testing whether two SQL queries are equivalent is computationally undecidable (or hard in the case of set semantics), and we therefore require approximations for testing SQL equivalence. Related work in Text2SQL evaluation has proposed such approximations, which they call \textit{match functions}, falling into two categories. Match functions in the first category, called \textit{semantic match functions}, define a set of (algebraic) rules to normalize two queries into an intermediate format, typically an abstract syntax tree (AST) format, and then check whether these two normalized queries are equivalent by comparing the intermediate format (e.g., via tree isomorphism testing). The second category, \textit{execution match functions}, compares the output relation of executing the SQL queries on database instance(s) to determine if the two queries are identical. Note that match functions in either category come with specific, input query-independent design choices. For instance, whether or not one should ignore the order in the output depends on the SQL queries; e.g., only queries with an \sql{ORDER BY} clause should consider the order. An additional, presumably more important, challenge when relying on existing match functions is that they typically encode a disambiguation process in the matching algorithm. For instance, if an execution match function simply ignores duplicates in the output, it implicitly assumes that, independently of the NL description, all \sql{SELECT ...} and \sql{SELECT DISTINCT ...} queries are equally likely. This can introduce a significant bias when measuring the performance of a specific sub-population of input samples, e.g., for distinct and non-distinct queries that are not equally likely.

As a result of ignoring both aforementioned challenges, trusting aggregate numbers in public benchmarks as a performance metric for generalization capabilities to use cases in practice can be misleading. On the one hand, a performance gap in a subgroup does not necessarily mean the model quality is poor but could be attributed to the inherent noise or bias in the evaluation process. On the other hand, a model achieving good performance might fail to generalize to production use cases if it overfits the noise or suffers from a large measurement bias.

\paragraph{A Taxonomy over Text2SQL Limitations}

In this work, we motivate and survey Text2SQL limitations yielding both prediction and evaluation errors using a new taxonomy. This taxonomy was built based on manual inspections of failures observed in state-of-the-art Large Language Models (LLMs) when solving Text2SQL tasks on public benchmarks and proprietary data. The primary purpose of this work is to raise awareness of all possible issues and limitations along with some real-world examples from state-of-the-art Text2SQL solutions and Spider, the most prominent public benchmark. We further offer suggestions on how these issues can be mitigated in practice. The secondary purpose of this work is to highlight the challenges in both practically applying these mitigation strategies and automatically performing coarse- or fine-grained categorization based on the taxonomy for a given Text2SQL solution. This is necessary to assess the magnitude of each issue in real-world use cases. However, the taxonomy cannot be directly applied to existing benchmarks due to several factors: (1) there currently exist no methods to detect and automatically categorize data quality issues, (2) prediction issues are dependent on the specific Text2SQL solution (i.e., the model trained), and (3) evaluation metric biases depend on the exact hyperparameters chosen and the predictions made. Many of the open challenges highlight the need for extensive future research in this area.

\section{Background}
\label{sec:background}

We formally introduce the Text2SQL task, describe aspects of LLM-based Text2SQL solutions, which currently achieve state-of-the-art results on public benchmarks, and conclude by describing the evaluation procedure adopted by these benchmarks.

\subsection{Text2SQL Task}

\begin{figure}
  \begin{sqllist}
  CREATE TABLE "Person" (
    "id"        INTEGER,
    "firstname" VARCHAR(100) NOT NULL,
    "lastname"  VARCHAR(100) NOT NULL,
    "age"       INT,
    "active"    CHAR(1) NOT NULL DEFAULT "T",
    PRIMARY KEY("id") )
  \end{sqllist}
  \vspace{-1em}
  \caption{Example schema}
  \label{fig:example_schema}
  \vspace{-1em}
\end{figure}

As is typical in machine learning, we define a \textit{task} over an input and output space. For a fixed schema, we define the \textit{input space} $\mathcal{X}$ as tuples consisting of two strings: (1) a natural language description and (2) a serialized version of the schema. The \textit{output space} $\mathcal{Y}$ contains all executable, i.e., valid SQL queries for the given schema.
The Text2SQL task is to predict exactly one, i.e., the most likely or any of the most likely, valid, and executable SQL query matching the natural language description for the given schema.

\begin{table}[t]
\small
\begin{tabular}{@{}rl@{}}
\toprule
Nr. & SQL Query                                                      \\ \midrule
    & NL1: ``List the names of all the people''                         \\ \midrule
Q1.1  & \verb~SELECT firstname, lastname FROM Person;~                      \\
Q1.2  & \verb~SELECT firstname FROM Person;~                                \\
Q1.3  & \verb~SELECT firstname || " " || lastname FROM Person;~             \\ \midrule
    & NL2: ``Get me the first names of inactive people''                 \\ \midrule
Q2.1  & \verb~SELECT firstname FROM Person WHERE active != 'T';~            \\
Q2.2  & \verb~SELECT firstname FROM Person WHERE active = 'F';~             \\
Q2.3  & \verb~SELECT firstname FROM Person WHERE active != 'Y';~            \\
Q2.4  & \verb~SELECT firstname FROM Person WHERE active = 'N';~             \\ \midrule
    & NL3: ``What is the first name of the oldest person?''              \\ \midrule
Q3.1  & \begin{tabular}[c]{@{}l@{}}\verb~SELECT firstname FROM Person~\\ \verb~  WHERE age in (SELECT MAX(age) FROM Person);~\end{tabular}       \\
Q3.2  & \begin{tabular}[c]{@{}l@{}}\verb~SELECT firstname FROM Person~\\ \verb~  ORDER BY age DESC LIMIT 1;~\end{tabular}      \\
Q3.3 & \begin{tabular}[c]{@{}l@{}}\verb~SELECT firstname, age FROM Person~\\ \verb~  WHERE age in (SELECT MAX(age) FROM Person);~\end{tabular}  \\ \midrule
    & NL4: ``What ages do people have?''                                \\ \midrule
Q4.1 & \verb~SELECT DISTINCT age from Person;~                            \\
Q4.2 & \verb~SELECT DISTINCT age from Person WHERE age IS NOT NULL;~       \\
Q4.3 & \verb~SELECT age FROM Person;~                                     \\
Q4.4 & \verb~SELECT age FROM Person WHERE age IS NOT NULL;~              \\
Q4.5 & \verb~SELECT age, COUNT(*) FROM Person GROUP BY age;~             \\
Q4.6 & \begin{tabular}[c]{@{}l@{}}\verb~SELECT age, COUNT(*) FROM Person WHERE age IS NOT NULL~\\ \verb~  GROUP BY age;~\end{tabular}     \\ \bottomrule
\end{tabular}
\caption{Different possible queries for schema in Figure~\ref{fig:example_schema} and four different natural language descriptions.}
\label{tbl:distribution_examples}
\vspace{-3em}
\end{table}

\paragraph{Joint Input-Output Distribution}
To formally understand the impact of ambiguity, let us define an unknown \textit{joint distribution} over the input and output space. Formally, let $X$ and $Y$ be two random variables taking values in $\mathcal{X}$ and $\mathcal{Y}$, respectively. We define the joint distribution $p(x,y) = p(X=x, Y=y)$. The conditional distribution $p(Y \vert X)$ represents the \textit{generative process} in Text2SQL, which infers $y$ for a fixed $x$ by sampling from this distribution.

When inspecting the conditional distribution $p(Y \vert X)$, we see that due to ambiguity in natural language and missing information in the serialization process of the schema, or more generally, the fact that there might exist many different valid SQL queries with non-zero probability, there is typically not a single ground truth SQL query for a fixed input. Formally, the set $\argmax_{y\in \mathcal{Y}} p(y \vert x)$ can contain more than one element, or more generally, $\max_y p(y \vert x)$ is smaller than 1 for these inputs $x$. We provide some example queries in Table~\ref{tbl:distribution_examples}, which are different yet all plausible interpretations of the provided NL description and schema in Figure~\ref{fig:example_schema}.

\paragraph{LLM-Based Solutions}

LLM-based solutions have been at the forefront of recent progress in Text2SQL~\citep{li2024codes, rajkumar2022evaluating}. There are typically three steps involved when using pre-trained or fine-tuned LLMs to solve the Text2SQL task:

\begin{enumerate}[leftmargin=*]
    \item Input preparation: Build the actual prompt based on the natural language description and the serialized schema. This can be a zero- or few-shot approach and, if the serialized schema is too large for the prompt length, might involve retrieving only the relevant tables from the schema.
    \item Run inference: Get the completion result based on next-token predictions and some hyperparameters.
    \item Output extraction: Extract the SQL query from a decoded text prediction and try executing it against an example database instance for the schema to catch syntax errors or timeouts.
\end{enumerate}

\subsection{Text2SQL Evaluation}

To programmatically evaluate a Text2SQL solution and assess the quality of the predicted SQL queries, benchmarks use an unseen evaluation or test dataset along with an automatic metric to compute the average accuracy over all predictions in this evaluation dataset. Given that the task requires an SQL query to be predicted by the Text2SQL solution, automatic metrics typically mark samples where the solution does not yield any executable prediction as incorrect. For executable predictions, metrics typically compare a single ground truth (GT) label provided per input in benchmark datasets like Spider~\citep{yu2018spider} or BIRD~\citep{li2024can} against the predicted SQL query. Note that an ideal metric would check for SQL equivalence, a well-defined term in database literature~\citep{abiteboul1995foundations}. The caveat with testing for SQL equivalence is that it has been proven computationally hard or even undecidable in general~\citep{abiteboul1995foundations}. Evaluating Text2SQL solutions, therefore, requires practical approximations to SQL equivalence, which are called match functions (MFs) in the literature.  Most prominent MFs can be divided into two categories: (1) semantic match functions, which compare the SQL queries by applying valid algebraic transformations, and (2) execution match functions, which compare the results of executing both queries (GT and prediction) on valid DB instance(s) for the given schema. As we will show later in this paper, either approach can lead to different under- or over-estimation issues. An additional property of these match functions is that, given that there is only a single GT label provided in the evaluation datasets, they typically include some relaxations to SQL equivalence to capture ambiguity and compare against different, yet equally likely, queries.

\begin{table}[t]
\begin{tabular}{@{}lll@{}}
\toprule
Error                             & Sub-Type           & Causes (Limitations)                 \\
\midrule
\multirow{2}{*}{Prediction} & Missing prediction & Solution or Eval Data \\
                            & Wrong prediction   & Solution or Eval Data \\
\multirow{2}{*}{Evaluation} & Type I error (FP)  & Eval Data or Eval Metric \\
                            & Type II error (FN) & Eval Data or Eval Metric \\ 
\bottomrule
\end{tabular}
\caption{Prediction and evaluation errors with their causes. We refer to `solution' as the full Text2SQL solution to be evaluated (e.g., an LLM with the prompt building and result extraction strategy).}.
\label{tbl:errors}
\vspace{-3em}
\end{table}

\section{Taxonomy of Text2SQL Limitations}

To motivate our taxonomy of Text2SQL limitations, we start by defining prediction and evaluation errors, along with their potential causes (cf. Table~\ref{tbl:errors}).

\begin{table*}[]
\begin{tabular}{@{}lllll@{}}
\toprule
Limitation (1st level)             & Issue (2nd level)                                                                        & Issue (3rd level)                                                                            & Resulting Errors                                                                                    & Possible Mitigations                                                                                         \\ \midrule
\multirow{5}{*}{Text2SQL Solution} & \multirow{2}{*}{Input preparation}                                                       & Missing information                                                                          & \begin{tabular}[c]{@{}l@{}}Mostly wrong predictions, \\ sometimes missing predictions\end{tabular}  & \begin{tabular}[c]{@{}l@{}}Augment input / \\ test for prompt length\end{tabular}                            \\ \cmidrule(l){3-5} 
                                   &                                                                                          & Suboptional prompt                                                                           & \begin{tabular}[c]{@{}l@{}}Mostly wrong predictions,\\ sometimes missing predictions\end{tabular}   & Prompt engineering                                                                                           \\ \cmidrule(l){2-5} 
                                   & \multirow{2}{*}{Inference step}                                                          & API / System failures                                                                        & Missing predictions                                                                                 & \begin{tabular}[c]{@{}l@{}}Testing (APIs / prompt\\ length / safeguards)\end{tabular}                        \\ \cmidrule(l){3-5} 
                                   &                                                                                          & Model misprediction                                                                             & \begin{tabular}[c]{@{}l@{}}Missing predictions,\\ or wrong predictions\end{tabular}                 & \begin{tabular}[c]{@{}l@{}}Continue training /\\ change of architecture\end{tabular}                         \\ \cmidrule(l){2-5} 
                                   & Result extraction                                                                        & n/a                                                                                          & \begin{tabular}[c]{@{}l@{}}Mostly missing predictions,\\ sometimes wrong predictions\end{tabular}   & \begin{tabular}[c]{@{}l@{}}Constrained decoding / \\ fine-tune model\end{tabular}                            \\ \midrule
Evaluation Data                    & Label accuracy                                                                           & n/a                                                                                          & Type I or Type II eval errors                                                                       & Clean labels                                                                                                 \\ \cmidrule(l){2-5} 
                                   & \multirow{4}{*}{Label completeness}                                                      & Distinct vs. not-distinct                                                                    & Type II eval errors                                                                                 & \begin{tabular}[c]{@{}l@{}}Change task /\\ Augment labels\end{tabular}                                       \\ \cmidrule(l){3-5} 
                                   &                                                                                          & Projection clauses                                                                           & Type II eval errors                                                                                 & \begin{tabular}[c]{@{}l@{}}Change task /\\ Augment labels\end{tabular}                                       \\ \cmidrule(l){3-5} 
                                   &                                                                                          & Filter conditions                                                                            & Type II eval errors                                                                                 & \begin{tabular}[c]{@{}l@{}}Change task /\\ Augment labels\end{tabular}                                       \\ \cmidrule(l){3-5} 
                                   &                                                                                          & Schema ambiguity                                                                             & Type II eval errors                                                                                 & \begin{tabular}[c]{@{}l@{}}Change task /\\ Augment labels\end{tabular}                                       \\ \cmidrule(l){2-5} 
                                   & \multirow{2}{*}{Feature accuracy}                                                        & Unanswerable inputs                                                                          & Wrong predictions                                                                                   & \begin{tabular}[c]{@{}l@{}}Change task /\\ Remove sample\end{tabular}                                        \\ \cmidrule(l){3-5} 
                                   &                                                                                          & Wrong presuppositions                                                                              & Wrong predictions                                                                                   & \begin{tabular}[c]{@{}l@{}}Change task /\\ fix presupposition\end{tabular}                                     \\ \cmidrule(l){2-5} 
                                   & Feature completeness                                                                     & n/a                                                                                          & Wrong or missing predictions                                                                        & Augment features                                                                                             \\ \midrule
\multirow{8}{*}{Evaluation Metric} & \multirow{2}{*}{\begin{tabular}[c]{@{}l@{}}SQL equivalence\\ approximation\end{tabular}} & Semantic match                                                                               & Type II eval errors                                                                                 & \begin{tabular}[c]{@{}l@{}}Test for more valid\\ algebraic transformations\end{tabular}                      \\ \cmidrule(l){3-5} 
                                   &                                                                                          & Execution match                                                                              & \begin{tabular}[c]{@{}l@{}}Mostly type I eval errors, \\ sometimes type II eval errors\end{tabular} & \begin{tabular}[c]{@{}l@{}}Test other DB instances\\ (especially when yielding\\ empty results)\end{tabular} \\ \cmidrule(l){2-5} 
                                   & \multirow{6}{*}{Ambiguity relaxation}                                                    & Ignore row order                                                                             & \begin{tabular}[c]{@{}l@{}}Missing: Type II eval errors\\ Applied: Type I eval errors\end{tabular}  & n/a                                                                                                          \\ \cmidrule(l){3-5} 
                                   &                                                                                          & Ignore duplicate rows                                                                        & \begin{tabular}[c]{@{}l@{}}Missing: Type II eval errors\\ Applied: Type I eval errors\end{tabular}  & n/a                                                                                                          \\ \cmidrule(l){3-5} 
                                   &                                                                                          & Ignore result types                                                                          & \begin{tabular}[c]{@{}l@{}}Missing: Type II eval errors\\ Applied: Type I eval errors\end{tabular}  & n/a                                                                                                          \\ \cmidrule(l){3-5} 
                                   &                                                                                          & Ignore column order                                                                          & \begin{tabular}[c]{@{}l@{}}Missing: Type II eval errors\\ Applied: Type I eval errors\end{tabular}  & n/a                                                                                                          \\ \cmidrule(l){3-5} 
                                   &                                                                                          & \begin{tabular}[c]{@{}l@{}}Flatten relation into list or\\ set (ignore columns)\end{tabular} & \begin{tabular}[c]{@{}l@{}}Missing: Type II eval errors\\ Applied: Type I eval errors\end{tabular}  & n/a                                                                                                          \\ \cmidrule(l){3-5} 
                                   &                                                                                          & Test for overlap only                                                                        & \begin{tabular}[c]{@{}l@{}}Missing: Type II eval errors\\ Applied: Type I eval errors\end{tabular}  & n/a                                                                                                          \\ \bottomrule
\end{tabular}
\vspace{1em}
\caption{Full taxonomy of Text2SQL limitations resulting in prediction or evaluation errors.}
\label{tbl:taxonomy}
\end{table*}

\paragraph{Prediction Errors}
There are two types of prediction errors for a fixed input: (a) a missing (i.e., no executable) query when there exists at least one valid SQL query for the input (i.e., NL and schema), and (b) a wrong prediction, for which executing the SQL query in \textit{any} valid database instance for the schema results in missing tuples, wrong tuples, missing attributes, or wrong attributes when comparing the result of the predicted SQL with \textit{any} valid SQL interpretation of the natural language over the schema. While the major cause for either missing or wrong predictions can often be found in limitations of the Text2SQL solution, they can also be a result of noise in the evaluation data. For instance, if the input features do not contain all the schema-specific information beyond common-sense knowledge required to translate a natural language description into a specific SQL query, failing to predict this SQL query is not necessarily (only) a limitation of the Text2SQL solution but can rather be caused by these missing features or a combination of evaluation data and Text2SQL solution limitations.

\paragraph{Evaluation Errors}

Complementary to prediction errors, we observe errors in assessing the quality of a proposed Text2SQL solution when computing the accuracy of predictions over a test set. Following traditional terms of binary classification and statistical hypothesis testing, we see two types of errors: (a) type I errors or False Positives (FP), when the evaluation process wrongly classifies a prediction for a fixed input as `correct', and (b) type II errors or False Negatives (FN), when the evaluation process wrongly classifies a prediction for a fixed input as `incorrect'. Both types of errors can be caused by either bias in the evaluation metric or noise in the evaluation data, more specifically the provided ground truth labels.

\paragraph{Unanswerable or Ambiguous Queries} The main causes for noise in the evaluation data or bias in the evaluation metric are unanswerable or ambiguous queries. We will point to such queries from different angles while outlining the limitations based on our taxonomy. In this work, we say a query is ambiguous when there are multiple equally likely interpretations of an NL description and a schema. Conversely, a query is unanswerable if it is not related to the schema or does not have any plausible answer due to wrong presuppositions. A prominent example of wrong presuppositions is represented by top-k queries like NL3 in Table~\ref{tbl:distribution_examples}. If more than $k$ elements meet the filter conditions, which based on the phrasing of the questions is assumed not to be the case, the question is not answerable given this wrong presupposition. Q3.1 and Q3.2 give the same result if there is only one oldest person in the database. But what should the behavior of the Text2SQL solution be if there is more than one oldest person? We see three different alternatives: the system returns all tuples (i.e., Q3.1), ignoring the singular part of the instructions in the NL description; it returns a single oldest person at random (i.e., Q3.2), which is only one possible answer; or it should first check if there is a single oldest person and, only if so, return the corresponding query, which ultimately represents a different task before solving the Text2SQL task. The `right' behavior should be encoded in the instruction and used in the evaluation accordingly, as we will discuss in Section~\ref{sec:eval_data}.

\paragraph{A Taxonomy over Common Limitations} 
We observe three different high-level limitations resulting in both prediction and evaluation errors: (1) Text2SQL Solution Limitations, (2) Evaluation Data Limitations, and (3) Evaluation Metrics Limitations. These limitations form the natural top-level categories in our taxonomy, which was constructed based on manual inspections of failure cases observed in state-of-the-art LLMs addressing Text2SQL tasks across public benchmarks like Spider and proprietary datasets. For Text2SQL Solution Limitations (1), these are independent of the evaluation process and pertain solely to the model's ability to interpret the input and generate accurate SQL queries, resulting in prediction errors. In contrast, Evaluation Data Limitations (2) can affect both prediction and evaluation outcomes due to issues such as poorly annotated examples, ambiguous queries, or schema mismatches, which introduce inconsistencies and challenges in assessing model quality. Lastly, Evaluation Metrics Limitations (3) emerge from biases or deficiencies in the evaluation metrics themselves, making them independent of the predictions but capable of causing evaluation errors for a given Text2SQL solution.

The taxonomy highlights these categories while noting several challenges in its practical application to existing benchmarks. Firstly, there currently exist no methods for detecting and automatically categorizing data quality issues within these benchmarks. Secondly, prediction-related issues are inherently tied to the specific Text2SQL solution employed, making it challenging to generalize across different models or configurations. Lastly, biases in evaluation metrics can vary based on hyperparameter choices and the predictions made, further complicating objective assessments. Table~\ref{tbl:taxonomy} provides an overview of the taxonomy, which organizes these high-level limitations into subcategories to offer a comprehensive framework for analysis. The next three sections delve into these limitations, exploring causes and presenting real-world examples from public benchmarks and state-of-the-art solutions, followed by a discussion of potential mitigation strategies and open challenges.

\section{Text2SQL Solution Limitations}
\label{sec:text2sql_solution}

As written in Section~\ref{sec:background}, we focus on LLM-based Text2SQL solutions in our taxonomy, which typically follow a three-step approach: (1) input preparation, (2) running inference, and (3) output extraction. We organize the issues in our taxonomy accordingly.

\subsection{Input Preparation Issues}

We see two closely entangled reasons why the input might be ill-prepared: (a) losing relevant information in the input, and (b) following a sub-optimal prompt strategy. Both issues arise from the fact that the models used to translate natural language and a serialization of the schema have a limited context length in terms of tokens that can be input. Therefore, especially for large schemata, one has to filter out irrelevant information and limit the number of additional tokens used as instructions to the ML model. Both of these issues can lead to wrong or missing predictions. Such an example can be found in the test database \sql{university\_rank} and the natural language description "What is the total enrollment of universities with a overall rank 5 or below?". Most state-of-the-art LLMs predict a SQL query with a filter condition \sql{rank <= 5}, which could be correct for some interpretation of rank, yet in this context, the right filter condition is \sql{rank >= 5}, meaning that the semantics of a rank in this context are unclear and should be fixed as part of the input. Note that this issue could additionally (or instead) be assigned to a prediction issue of the Text2SQL solution if one assumes this is common knowledge, or to evaluation data issues if the semantics is supposed to be part of the schema and thus encoded in its serialized version.

\subsubsection{Mitigation Approaches}

The standard mitigation approach to add useful information to the prompt, proposed by the literature, is what has come to be called `Prompt engineering`~\citep{radford2019language, wei2022chain, sahoo2024systematic}. The caveat with this is that adding more and more information to a prompt might not be feasible due to context length limitations, or, if the context length is too large, it can hurt the model's ability to perform a prediction correctly~\citep{balachandran2024eureka}. To reduce irrelevant information about the schema, researchers have proposed applying retrieval-style algorithms over the tables~\citep{biswal2024text2sql, wu2024stark, chen2024table}. It is clear that to balance the size of the prompt and not miss out on relevant information, one must carefully find a good trade-off between the precision and recall of such a retrieval approach.

\subsubsection{Open Challenges}

Automatically detecting input preparation issues, and especially disentangling them from prediction issues, is still a largely under-explored area of research. To prevent information loss during retrieval and avoid not answering a query or giving a wrong (i.e., hallucinated) response, one might consider using a verification module as part of the Text2SQL solution~\citep{zhang2023towards, bayat2023fleek}, bearing in mind that this additional step might increase prompt length and latency.

\subsection{Inference Issues}

The simplest form of inference issues arises based on system errors like API failures (e.g., if the prompt is too long) or safeguard violations, resulting in missing predictions. The more complex form of inference issues is what related work has mainly focused on: model prediction issues. Note that such model prediction issues, often referred to as "hallucination" or lack of "reasoning capabilities," can either result in no executable SQL query if the prediction does not contain any SQL query or it has an invalid syntax (e.g., hallucinated table name or attribute), or a wrong SQL query if the produced SQL query is syntactically correct but yields an incorrect result on any valid database instance.

\subsubsection{Mitigation Approaches}

Handling system errors is fairly trivial via exception handling as part of a Text2SQL solution. To mitigate model prediction issues, one can, after having identified an underperforming sub-population (e.g., complex queries with some measure of complexity), try applying established techniques to close the performance gap for that sub-population, like model patching~\citep{goel2020model} or a change in model design or training procedure.

Another approach to mitigating prediction errors is to change the \textit{task} itself by asking the model (e.g., via prompt engineering or repeated sampling during the decoding process) to generate multiple different SQL queries for the fixed input features. This idea can not only resolve evaluation issues (i.e., when there are multiple different plausible interpretations) but also fix small semantic or syntactic errors.  Experimenting with different open LLMs, we have found that prompting the model to generate multiple variants helps resolve many of the prediction issues for simple queries, especially when relying on smaller models.

\subsubsection{Open Challenges}

Identifying the reasons why specific sub-populations or groups experience poor model performance is an under-explored area of research~\citep{rawte2023survey}. The main difficulty lies in its entanglement with input preparation issues, and more specifically, prompt strategies.  One example is given by \citet{abbe2024far}, where the authors propose to jointly change the prompt and training procedure to enhance certain LLM capabilities.

\subsection{Result Extraction Issues}

Even if the model manages to produce the `right' SQL query, extracting it from the decoded prediction in text form is not always straightforward. State-of-the-art models sometimes complicate the extraction process by fusing explanations into the query output using improper comments, relying on no or non-standard coding block syntax, or merging multiple coding blocks into one. Ultimately, the task of result extraction is to extract the full, valid SQL query from an unstructured textual prediction.  Achieving this task typically combines two things: (1) clarifying the desired output format as part of the prompt instructions (e.g., using the triple prime SQL coding block syntax), and (2) parsing based on that format and a custom regular expression (regex). A failure in this extraction process mostly leads to a missing prediction, but, depending on the input and extraction process, it can also parse a syntactically valid part of the full query and thus result in an incorrect prediction.

\subsubsection{Mitigation Approaches}

Any programmatic extraction algorithm should be tested via well-established software engineering best practices, such as unit testing. The standard approach to enforcing the desired output format in practice is to iterate on the instructions (e.g., via few-shot examples) in the prompt. This approach does not give any guarantees on the output though and, as outlined before, could lead to other issues due to an increase in prompt length. Alternatively, one can fine-tune a fixed model for a small number of iterations to bias the model toward the desired output format. Note that this bias could hurt the generalization performance of the model. Another solution to this problem is to change the decoding process to \textit{constraint decoding}, which allows only outputs following some fixed grammar~\citep{geng2023grammar}. The downside of this approach is that it is difficult to capture the full grammar across SQL and arbitrary schemata, and checking for the grammar during the decoding process can be computationally demanding, thus increasing latency.

\subsubsection{Open Challenges}

The main challenge in controlling the output format lies in balancing the accuracy of an extraction process with its latency.  On the bright side, our manual inspection revealed that for currently available benchmark datasets and state-of-the-art models, detecting result extraction failures is rather simple, given that most of the missing predictions for these datasets stem from result extraction errors only. While this is true for these relatively simple and answerable inputs, we conjecture that disentangling the causes of missing prediction errors among input preparation, model prediction, and evaluation data issues on real-world distributions will represent an important research direction for the future.

\section{Evaluation Data Limitations}
\label{sec:eval_data}

\begin{figure*}[ht]
    \centering
    \includegraphics[width=0.9\linewidth]{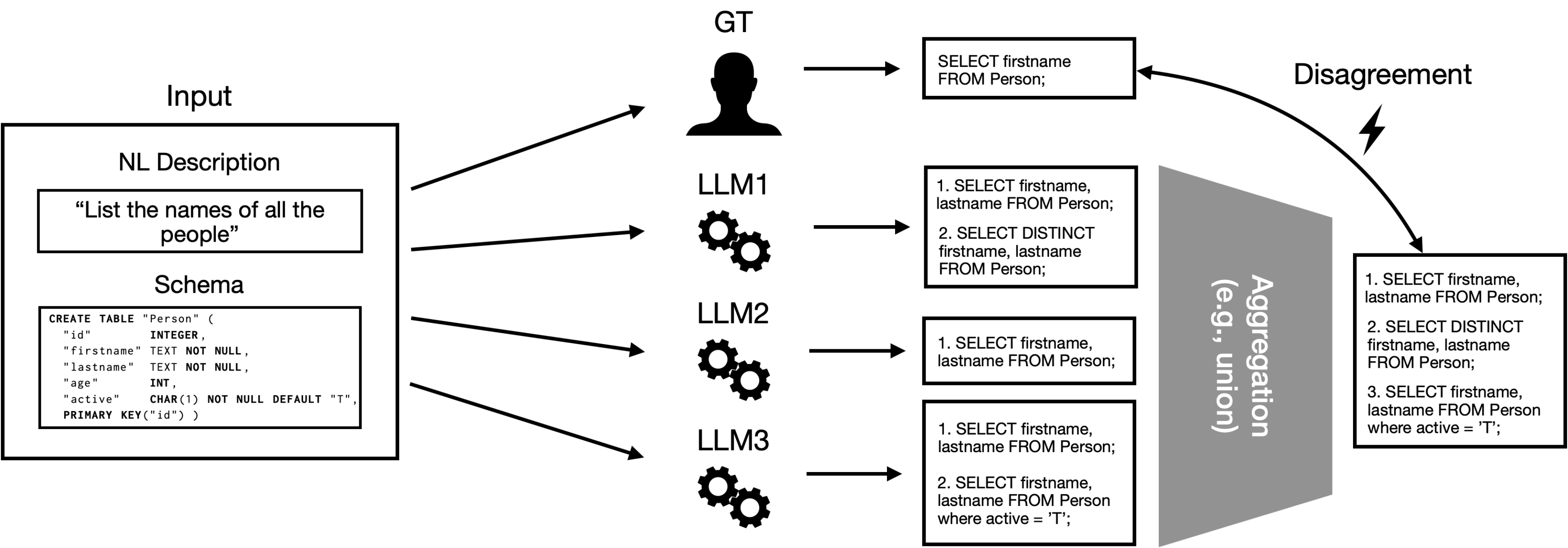}
    \caption{Noisy Data Detection: For a fixed input (NL description and serialized schema), we ask multiple LLMs as independent voters to predict the top 3 most likely SQL queries. The variants across all voters are kept and compared against the single ground truth query in the benchmark dataset.}
    \label{fig:llm_based_approach}
\end{figure*}

By construction, datasets used in Text2SQL benchmarks like Spider were designed such that they only contain high-quality data. They rely on skilled human annotators and review process ensuring that each question is answerable without ambiguity or requiring knowledge outside of the database (cf. Section 3 in \citet{yu2018spider}). While it might seem that this goal is achieved at first glance, we found many subtle data quality issues specific to the Text2SQL task in these datasets. These data issues introduce an irreducible error into the performance \textit{any} Text2SQL solution can achieve.

We categorize data quality issues for both labels and features along two dimensions classically used in the data management literature: accuracy and completeness~\citep{sidi2012data}. Note that in the general context of AI and ML, research is still in its early stages in understanding the effects of data quality issues, particularly concerning unstructured data such as images and text.~\citep{renggli2021data}. In this section, we will focus on Text2SQL-specific limitations, leaving a more general viewpoint for future work.

\subsection{Label Inaccuracy}

The most obvious evaluation data limitation is reflected by inaccurate labels. More specifically, the labels of samples in the evaluation dataset are inaccurate if the given GT label query is not executable, or its result is wrong or incomplete. While in general, a label could be completely wrong, we found that there are no non-executable queries, but multiple minor issues resulting in wrong SQL queries for high-quality datasets like Spider. For example, in the database \sql{book\_press}, one has to consider that the attribute \sql{sale\_amount} is represented as text, not an integer. The GT label
\begin{lstlisting}
    SELECT title, release_date FROM book ORDER BY sale_amount DESC LIMIT 5
\end{lstlisting}
for the NL description "What are the 5 best books in terms of sale amount? Give me their titles and release dates." will therefore yield a wrong result on the provided database instance. Another set of label inaccuracies concerns filter conditions. Some prominent examples include case-sensitive string comparisons or improper use of the `NULL' comparison. For instance, in \sql{movie\_2}, the NL description "What are the names of all cinemas not showing any movies?" is translated to
\begin{lstlisting}
    SELECT DISTINCT name FROM MovieTheaters WHERE Movie = 'null',
\end{lstlisting}
where the filter condition should be \sql{is NULL}. Some more subtle errors are due to improper interpretation of SQL semantics. For instance, in \sql{warehouse\_1}, the SQL query 
\begin{lstlisting}
    SELECT sum(T1.value) FROM boxes AS T1 JOIN warehouses AS T2 ON T1. warehouse = T2. code ORDER BY T2. capacity DESC LIMIT 1
\end{lstlisting} does not represent the NL description "Find the total value of boxes stored in the warehouse with the largest capacity." because the \sql{sum} operator is executed before the \sql{LIMIT} operator, thus computing the sum over all values in the database instead of only those for the warehouse with the largest capacity.
Another example can be found in the database \sql{book\_1}. The NL description "What are the titles of books that have a sale price equal to the lowest sale price across all books?" clarifies the presupposition issue for top-k queries mentioned earlier, yet its GT label
\begin{lstlisting}
    SELECT title FROM book ORDER BY saleprice ASC LIMIT 1
\end{lstlisting}
is clearly wrong in that it only returns a random element meeting the condition and should be fixed accordingly.

Overall, wrong labels tend to increase type II evaluation errors (FNs). In some cases, though, label inaccuracies can result in type I evaluation errors (FPs) if the ML model `randomly' made the same mistake as the GT labeler, or if it overfits to the evaluation set due to data leakage.

\subsubsection{Mitigation Approaches}

Although finding wrong labels is hard, which we will elaborate on in the next paragraph, fixing most of them is typically straightforward for an expert human labeler. For complex queries, one might be required to break them into pieces to find the actual mistake. One such example can be found in the database \sql{aan\_1} for the NL description "Find the number of authors who did not publish any paper that is cited more than 50 times.". The GT label
\begin{lstlisting}
    SELECT count(*) FROM Author WHERE Author_id NOT IN ( SELECT T2.author_id FROM Citation AS T1 JOIN Author_list AS T2 ON T1.cited_paper_id = T2.paper_id GROUP BY T1.cited_paper_id HAVING count ( DISTINCT T1.paper_id ) > 50 )
\end{lstlisting}
is wrong. When changing the query to return the author IDs instead of an aggregate count, we see that the author with ID `494' is wrongly returned. This author is on paper `N09-1003`, which is cited more than 50 times. This issue is very subtle and arises from the projection \sql{T2.author\_id} of the sub-select statement, an attribute to which the \sql{group by} operator was not applied. While, according to SQL standards, this query should result in an error, some RDBMS like SQLite execute this query but return an arbitrary value of \sql{T2.author\_id} from the grouped rows, which can result in wrong behavior. Fixing this requires rewriting the entire query using a different approach (e.g., using two sub-selects).


\subsubsection{Open Challenges}

For the databases in Spider, we have found that manually checking the labels of queries resulting in an \textit{empty result} (i.e., the same result as if the query is executed against an empty database) is a good proxy to find some of the wrong labels. While this approach does not generalize to benchmark datasets where there are queries resulting in empty results on purpose, we found that out of the 25 queries in the test set of Spider resulting in an empty result (roughly 1\% of the full dataset), 21 have a wrong label, mostly due to wrong value comparison or null handling, one is a correct interpretation of how to handle \sql{NULL} values, yet not the most plausible one, and the remaining three suffer from language ambiguity when constructing the filter conditions and are likely to represent wrong labels as well. For instance, in the database \sql{art\_1}, the NL description "What are the locations that have works painted before 1885 and after 1930?" and label
\begin{lstlisting}
    SELECT DISTINCT LOCATION FROM paintings WHERE YEAR < 1885
    INTERSECT
    SELECT DISTINCT LOCATION FROM paintings WHERE YEAR > 1930    
\end{lstlisting}
could be a valid interpretation, yet replacing \sql{intersect} with \sql{union} would make more sense, given that no work can be painted both before 1885 and after 1930.

In general, detecting wrong labels in a Text2SQL benchmark is a tedious and hard task, for which there is currently no benchmark to evaluate automatic approaches. Ideally, one would ask multiple independent human experts to check the labels, which is very costly and does not scale. As an approximation to this, we compare the disagreement cases between the GT label and the results of state-of-the-art LLMs prompted to return multiple plausible variants, as depicted in Figure~\ref{fig:llm_based_approach}. If the GT label did not agree among the different match functions and any of the proposed SQL variants (i.e., a union over all proposed variants), we manually checked them. Note that this approach is merely a proxy to find interesting samples and does not differentiate between different data quality issues. We provided multiple examples of inaccurate labels in Spider in this section. A detailed overview of the proxy method and an extended, yet not necessarily complete, list of manually annotated wrong labels for Spider discovered via the proposed LLM-based approach can be found in the supplementary material (cf., Appendix~\ref{app:evluation_data_issues} in the supplementary material). It is worth noting that in the context of structured input data (i.e., numerical or categorical values), research in data cleaning for ML has proposed methods to automatically detect and clean data~\citep{ilyas2022machine}. Similarly, for multi-class classification over unstructured data, various surveys and methods have been proposed to estimate the amount of label noise~\citep{northcutt2021pervasive, renggli2021evaluating}. Extending these methods to label-independent or GenAI use cases and evaluating them in the context of multi-modal inputs with unstructured data like text is an open challenge requiring future research.

\subsection{Label Incompleteness}

The second dimension captures the core issue with asking for a single GT SQL query for every possible input, especially for the NL part. NL is known to be inherently ambiguous. Together with possible database or schema issues, it is clear that -- even for supposedly non-ambiguous queries in Spider -- not all the ambiguity can be removed from the input. Therefore, we find many cases where different SQL queries are plausible interpretations for the fixed input. We divide them into four different sub-categories, based on the structure of SQL queries. Before outlining them, note that missing ambiguous variants for fixed inputs can typically result in FN evaluation errors.

\paragraph{1. Distinct vs. not-distinct} Without explicitly mentioning "unique" or "distinct" in the NL description, it is often unclear whether the SQL query should return all (i.e., including duplicates), or only distinct values. The same holds in the case of aggregations like \sql{count}, and whether they should be conducted over all or distinct values only. We see many cases for the same database, e.g., \sql{bakery\_1}, where similar questions have a \sql{distinct} clause in one provided GT label for a fixed NL description (e.g., for "What are the full names of customers who bought apple flavored Tarts?"), but not for another seemingly similar NL description (e.g., "What is the last name of the customers who shopped at the bakery more than 10 times?"). This phenomenon is even more present when comparing queries across databases and schemas.

\paragraph{2. Projection clauses} Without having very clear and detailed instructions in the NL description, it is often unclear exactly which attributes of the schema, and in what order, if relevant, are to be returned. For instance, in the database \sql{boat\_1}, the NL description "Find boats reserved by Sailor with id 1." might request the id of boats, their name, or even all the relevant properties of boats. The impact of missing ambiguous variants in the projection clauses is even harder to capture if one allows for transformations over attributes. For instance, the example asking for full names of customers from before could either request all the name attributes separately or, according to some local conventions, request a string concatenation over the first name and last name (e.g., \sql{firstname || " " || lastname} in SQL dialect).

\paragraph{3. Filter conditions} The most obvious cases of ambiguity when it comes to filter conditions represent string comparisons or filters based on some notion of semantic similarity. Many of these aspects for exact string comparison can be prevented if the schema is restricted to a fixed-sized domain per attribute, communicated to the ML model during the serialization process of the schema. Another issue lies in the interpretation of \sql{NULL} values. According to SQL standards, these values are assumed to be unknown. One such example can be found in \sql{art\_1}, where the NL description "Find the names and years of all sculptures that are not located in gallery 226." is translated into 
\begin{lstlisting}
    SELECT title, year FROM c WHERE LOCATION != 'Gallery 226'
\end{lstlisting}. Note that this query does not return any result, but when checking the database, we find one sculpture having a \sql{NULL} location. Whether or not it should be included is unclear from the NL description; hence, having a filter condition \sql{LOCATION != `Gallery 226' or LOCATION is NULL} represents another valid interpretation.

\paragraph{4. Schema ambiguity} The final bucket of ambiguity captures all other schema-specific aspects, potentially leading to completely different results. The first and foremost issue we see in all BIRD databases and some of the Spider databases is not having normalized schemas. This allows for different plausible queries that, if the unnormalized data violates consistency requirements, can lead to different execution results. For instance, in \sql{e\_commerce}, the GT query for the question "What is the product name and the color of the ordered items which have been shipped?" can be answered by joining \sql{products, order\_item, shipment\_items, shipments} (i.e., what the GT query does), or joining \sql{products, order\_item, order, shipments} instead, leading to a different result on the provided database instance.

\subsubsection{Mitigation Approaches}

The standard mitigation approach for resolving these ambiguity issues is to relax the matching function, allowing approximate similarity between two SQL queries. This can help capture some fraction of the ambiguity cases outlined above but it can lead to other issues as we will outline in the next Section.

A more systematic approach would be to not only propose a single GT label per input sample but \textit{all} possible variants. A prediction for a fixed Text2SQL solution would then be accurate if it matches \textit{any} of the variants. If one were to envisage changing the Text2SQL task to allow a model to predict multiple variants, as suggested earlier in this work, one would need to change the evaluation metric from exact accuracy per prediction (i.e., one or zero) to some metric checking for overlap (e.g., F1 score). There exist preliminary studies in that direction~\citep{bhaskar2023benchmarking, wang2023know}, yet larger-scale benchmarks to evaluate these mitigation approaches are missing. 

\subsubsection{Open Challenges}

The first challenge with the proposed mitigation approach is to provide \textit{all} possible variants as labels, which can be very expensive to obtain if there are exponentially many variants. One might envisage using some semi-automatic or programmatic methods to augment existing labels with valid variants. Second, comparing a prediction against exponentially many variants can increase the computational demand on the evaluation process. Finally, to the best of our knowledge, there are currently no methods to automatically find samples with missing alternative variants (i.e., uncovered ambiguity cases). While using ideas of data integration or fusion~\citep{dawid1979maximum, ratner2020snorkel}, like changing the aggregation function for the method proposed in Figure~\ref{fig:llm_based_approach} to a more complex voting scheme-based solution, could offer a potential solution, it has yet to be fully explored and evaluated via new benchmarks. Preliminary results for noisy samples falling into this category are provided in the supplementary material in Appendix~\ref{app:evluation_data_issues}.

\subsection{Feature Inaccuracy}

We define a feature, i.e., the input to the Text2SQL solution, as inaccurate if there is no plausible and valid SQL interpretation for the given NL description over the fixed schema. 
Forcing a prediction in such cases inevitably leads to errors.  Inaccuracy arises primarily from two sources:  unrelated questions or wrong presuppositions.
First, an unrelated question lacks a corresponding SQL query. For example, "What is John Doe's email address?" is unanswerable if the schema (Figure~\ref{fig:example_schema}) contains no email address information.
Second, wrong presuppositions can create inaccuracy.  This is especially prevalent in datasets like Spider, which frequently contain top-k queries. Consider, "When was the transcript issued for the student with the maximum loan value?" in Database \sql{cre\_Students\_Information\_Systems}. If multiple students share the maximum loan value, the query is ill-defined. The Text2SQL system could either assume that the question refers to \textit{any} student with maximum load and randomly select one of the transcripts (using \sql{ORDER BY amount\_of\_loan DESC LIMIT 1}), or ignore the singular term in the NL description and return \textit{all} transcripts for students with the maximum loan value (i.e., using \sql{amount\_of\_loan in ( SELECT MAX(amount\_of\_loan) FROM Student\_Loans )}).

\subsubsection{Mitigation Approaches}

The standard approach to handling inaccurate features in the evaluation set is to remove or filter out the affected samples.  Alternatively, the task itself can be modified.  This could involve allowing for "no answer" responses or accepting multiple equally valid interpretations of the input under different viewpoints of semantic interpretations.  Finally, the natural language instructions could be clarified or augmented. For instance, in cases of top-k queries, instructions could explicitly state that any subset of at most $k$ tuples should be accepted, with ties broken randomly if more than $k$ tuples satisfy the criteria.

\subsubsection{Open Challenges}

We see two major challenges with feature inaccuracies. Firstly, detecting samples where the features are inaccurate can be very challenging, and except for \citet{wang2023know}, which conducted a preliminary analysis on unanswerable queries, there is, to the best of our knowledge, no related work targeting this problem in a principled way. To find questions with wrong presuppositions, one might parse the provided ground truth SQL label to match a fixed template (e.g., ending with an \sql{ORDER BY xzy LIMIT k} clause) and manually check these samples. For Spider, there are 335 samples (roughly 15\% of the test set size) matching this template and potentially resulting in a wrong prediction. Manual inspection of random samples showed that only a very small fraction of these 335 samples clarify the semantics towards returning all samples meeting the selection requirements (e.g., all tuples where an attribute is equal to the maximum over this attribute). Secondly, catching wrong predictions for wrong presuppositions is non-trivial. On the one hand, presented variations to resolving a presupposition issue often yield the same execution results when evaluated on a single, typically small, database instance, like the ones provided in Spider. On the other hand, if, in the case of top-k queries, the presupposition are clarified such that any random subset of $k$ tuples meeting the requirements should be returned, there is currently no evaluation method able to capture this nondeterministic behavior (cf., Section~\ref{sec:eval_metric}).

\subsection{Feature Incompleteness}

Incomplete features in the evaluation data represent the last data quality issue.  This manifests primarily as missing schema information, such as undefined or incomplete attribute domains and missing integrity constraints.  Consider all queries performing a value comparison, e.g., in the database \sql{bakery\_1} with the GT SQL query \sql{SELECT id, flavor FROM goods WHERE food = `Cake' ORDER BY price DESC LIMIT 1} and the corresponding NL description "What is the most expensive cake and its flavor?". Omitting the domain of the attribute \textit{food} (in this case containing the capitalized word "Cake") in the input to the Text2SQL solution typically results in a filter condition \sql{food = `cake'}. This condition is wrong for the given database but not necessarily for other valid database instances following the schema. To understand the impact of missing integrity constraints, let us inspect the database \sql{pilot\_1}. Based on the PK in the table \sql{Hangar}, one can assume that planes have unique names. Extrapolating this line of reasoning to pilots, one might assume that pilots are also uniquely identified by their names. Yet, depending on the meaning of the attribute \sql{age} in the table \sql{PilotSkills}, it might make more sense that a pilot is uniquely identified by their name \textit{and} age. How would one then differentiate between two pilots of the same age and a single pilot owning more than one different plane? These missing integrity constraints can lead to different valid SQL queries for most NL descriptions given for that database.

\subsubsection{Mitigation Approaches}

The straightforward approach to mitigate these issues is to (1) add all necessary schema information to the database by specifying domains for attributes and all valid integrity constraints, and (2) serialize this information when preparing the input for the Text2SQL solutions. Another solution to bypass the necessity of putting exact database values into filter conditions is to extend the RDBMS with semantic operators to filter tuples based on semantic similarity on some attributes~\citep{patel2024lotus}.

\subsubsection{Open Challenges}

Including the entire relational schema in the prompt for the Text2SQL model presents a significant challenge: the serialized schema can easily exceed the database size, creating an excessively large prompt. This, as discussed in Section~\ref{sec:text2sql_solution}, can ultimately increase the amount of prediction errors. To address this, we could augment the input with only the relevant missing schema parts, perhaps using random database samples. However, identifying which samples are affected by missing schema components remains an open problem.

\section{Evaluation Metric Limitations}
\label{sec:eval_metric}

There are two main issues that can lead to evaluation metric limitations. First, testing for SQL equivalence is computationally hard and requires practical approximations. Second, given that there is only a single ground truth SQL query label and inherent ambiguity in the natural language descriptions, researchers have incorporated relaxations into the process of comparing SQL queries, which can, in turn, lead to bias in the evaluation.
Understanding the impact of either issue is hard in practice, given that they depend on the input distribution and also on the output of the trained models. Deriving practical estimations of the bias impact over these input-output distributions remains an unsolved challenge.

\subsection{SQL Equivalence Approximation Issues}

There are typically two ways of approximating SQL equivalence: (1) comparison of two SQL queries at a semantic level without executing them, and (2) executing the two SQL queries on some database instance and comparing their results. Both approaches suffer from different approximation issues and result in different evaluation errors.

\paragraph{Semantic match} Approaches falling into this category typically first parse both a (string) GT and a predicted query into some intermediate format like an abstract syntax tree (AST). By applying valid algebraic transformations for the schema, they then test whether one of these ASTs can be transformed into the other. If this succeeds, the two SQL queries are defined as equivalent. For this approach to succeed, one requires a complete set of algebraic transformations and potentially a very large number of repeated applications of any selection of valid transformations. Thus, many of the proposed semantic match functions result in type II evaluation errors (FNs) if valid transformations are missing or the search process was terminated prematurely.

\paragraph{Execution match} These match functions use one (or multiple) valid database instances for the given schema. By executing two queries on the same DB instance, they can compare the execution results to assess the equivalence between two SQL queries. Note that, by definition, two queries are equivalent if they yield the same result on \textit{all} valid DB instances for the schema. With a limited number of DB instances to test on (one for both Spider and BIRD), many queries yielding the same result can be different on other DB instances. This is especially relevant for queries yielding empty results, where two very different queries can both yield empty results on a single DB instance. Testing on a limited number of DB instances is also problematic for inputs with wrong presuppositions where, as described above, multiple approaches to resolving the presuppositions issues lead to the same result when tested only on a single small database. Overall, these errors typically result in type I evaluation errors (FPs).

Another issue with testing based on execution results can be found in non-deterministic questions, where the user specifically asks for a random result among the candidates, e.g., "any of the oldest person", which is translated into \sql{ORDER by age DESC LIMIT 1}. A proper execution match function would compare against all possible candidate results and only mark the predicted query as correct if any of the results are equivalent. Comparing against a single candidate result can increase the FN rate.

\subsubsection{Mitigation Approaches}

There exist extensions to the default match functions deployed in Spider and BIRD for both approaches to approximating SQL equivalence. ESM+~\citep{ascoli2024esm+} extends semantic matching with more rules that include schema information like integrity constraints. For execution matching, Testsuits~\citep{zhong2020semantic} augments every database with at least four other carefully created variants to achieve larger coverage. Finally, Dr.Spider~\citep{chang2023drspider} uses a more adversarial approach to change the samples (feature or label), or part of the DB instance, to test for robustness.

\subsubsection{Open Challenges}

Efficiently testing for SQL equivalence has been a decades-old problem in the database management community~\citep{chu2018axiomatic}. While the related work mentioned in the previous section identified issues with existing benchmark functions, it does not manage to rigorously address them for any input-output distribution, and it is relatively easy to manually construct examples for which the type I and type II evaluation errors are still relatively high. Having a proxy measure for how large these evaluation errors are for a fixed input-output distribution remains an open problem. Using modern LLMs~\citep{zhao2023llm} as part of the evaluation process and extending them to incorporate human signals can offer interesting research opportunities~\citep{shankar2024validates} in this area.

\subsection{Ambiguity Relaxations Issues}

Exactly comparing a predicted SQL query to a single ground-truth query omits the inherent ambiguity of the problem.  Existing methods address this ambiguity using different relaxations in their query comparison functions. Note that Spider and BIRD do not have the same relaxations applied for testing for equivalence based on execution match. Spider's match function parses the query and maps the columns to schema attributes to ignore column order, whereas BIRD performs a set comparison over all the values, thus ignoring duplicates, column order, and column affiliation. In general, applying any relaxation can increase the FP rate, whereas omitting it can increase the FN rate. The magnitude of these errors depends on both the sample distribution and the Text2SQL solution.

\paragraph{1. Ignore row order} A standard relaxation typically applied when using execution match is to ignore the order of the results. This relaxation makes sense, in that SQL does not give guarantees on the row order except if a query asks specifically for it (i.e., using the \sql{ORDER BY} operator). When filtering Spider for samples where the GT query has an \sql{ORDER BY} operator, yet no \sql{LIMIT} operator, we see that 24 samples (roughly 1\% of the test dataset) represent such cases, where, if a Text2SQL solution has a wrong order of the tuples, it would result in FP evaluation errors.

\paragraph{2. Ignore duplicate rows} Ignoring duplicate rows is identical to omitting the \sql{DISTINCT} keyword in the final projection of a SQL query. When checking for simple projections vs. projections with distinct samples in the test set of Spider, we see that 521 (~24\%) of the queries have no distinct keyword, and 151 (~7\%) have a distinct keyword. Manual inspection showed that only a very small fraction of the 151 samples specifically ask for unique or distinct results. Therefore, for a large number of these samples, predicting a SQL query \textit{without} a distinct keyword, assuming no other label noise, might result in FP evaluation errors.

\paragraph{3. Ignore column type} When comparing execution results or testing for semantic equivalence, one might want to ignore column types; i.e., if an attribute is cast into another type, say from a string to a float. This relaxation typically makes sense when applied to schemata where attributes are not strongly typed (e.g., floats, ints, or dates stored as strings). The caveat with performing type-casting is that depending on the RDBMS, casting into arbitrary types might not fail but result in a different result, which, when compared against each other, can increase the FP rate for some predictions.

\paragraph{4. Ignore column order} Whether or not the order of the columns returned matters is typically a design choice when defining the match function if not specified in the NL description. Note that in Spider, there are many samples for which the projection order in the GT SQL query does \textit{not} match the order in which they appear in the NL description. 
If the system intent is to always follow the order of the attributes in the NL description, we found that Spider suffers from another large fraction of additional inaccurate labels, or, if the relaxation is applied, it can result in many false positives in the evaluation process. Another prominent example is represented by aggregates over group-by queries. The default convention taught in many DB courses is to return the group by condition as the first attribute in the result and the computed aggregation as a second attribute. This convention is not always followed in the GT labels in Spider. It is therefore natural for many Text2SQL solutions to predict another order of the columns compared to the GT labels. Note that checking for all possible column orders can be computationally demanding (i.e., with exponential complexity).

\paragraph{5. Ignore column affiliation} An alternative to bypass the computational requirement of testing for all possible column orders is to flatten the resulting relation into a list (or a set) of values and perform a comparison of this flattened data structure. Performing such a relaxation can increase the FP rate on queries where the same data is used in multiple different columns (e.g., asking for object details that are linked to each other). A prominent example for this include cyclic joins.

\paragraph{6. Ignore exact column overlap} It is not always obvious which columns have to be returned for a given NL description. For instance, the GT SQL query for the NL description "What is the name of the oldest painting and where is it located?" in \sql{art\_1} returns the properties \sql{title, location, year}, where one could argue that \sql{year} was not requested. But when looking at the NL description "Which state has most number of students?" in \sql{address\_1}, we see that the GT SQL query only returns the \sql{state}, not including the number of students. Should predictions be omitting \sql{year} in the first example, or adding the number of students in the second example, both be marked as wrong? A possible relaxation to this is to check whether the result of one query is equivalent to a subset over columns of the second result (ignoring empty results).

\subsubsection{Mitigation Approaches}

The problem with resolving ambiguity via match function relaxation is that these relaxations are typically input-independent and are applied to \textit{all} GT SQL queries and predictions. A first mitigation approach would be to decide whether to apply a relaxation depending on the input sample (possibly using the GT label or some query plan of the RDBMS). This would, for instance, enable filtering out order-dependent or order-independent results. In the same spirit, one could use some schema or RDBMS information through a query plan to perform better column order matching or overlapping tests. This approach was also proposed by \citet{floratou2024nl2sql}, yet it comes with the challenge that not all schema information, nor the query plans, are necessarily available to the Text2SQL evaluation system. Finally, one could augment the single GT label with additional variants to avoid having to rely on \textit{some} of the relaxations if they increase the FP rate by too much. If, for any of the relaxations outlined above, there is no significant increase in the FP rate when applied, we can safely keep it applied and do not have to augment the labels for all the affected inputs.

\subsubsection{Open Challenges}

The main challenge we see with ambiguity relaxations is that (a) they are often not clearly stated in the public benchmarks but have to be reverse-engineered via code inspection, and (b) the impact of the relaxations in terms of overestimation (FP rate) is not only dependent on the predictions of the proposed Text2SQL solution but also the input-output distribution, which can be very different for production use-cases compared to the provided benchmark datasets. We currently lack methodologies and tools to systematically evaluate the impact of these programmatic relaxations on any target distribution. One potential approach towards such a methodology is to follow a carefully crafted voting- or disagreement-based algorithm (similar to the one depicted in Figure~\ref{fig:llm_based_approach}) to identify the number of samples resulting in different predictions when applying or omitting a relaxation. We leave this idea for future work.

\section{Conclusions}

In this paper, we propose and motivate a taxonomy to classify limitations when building or evaluating Text2SQL solutions. We highlight the causes, mitigation approaches, and open challenges across three high-level limitations: (I) limitations of the solution itself resulting in wrong or missing predictions, (II) limitations with the evaluation data resulting in either wrong or missing predictions if the features are corrupted, or over- or under-estimation of the quality of the model if the labels are corrupted, and (III) limitations of the evaluation process and the used metric resulting in over- or under-estimation of the model quality. Our work takes a first step towards building a complete taxonomy of prediction and evaluation issues for Text2SQL systems, being aware that it might miss some relevant categories and could be extended in the future.

\paragraph{Cross Top-Level Limitation Mitigation}

We see that mitigating a single issue can quickly result in an increase in issues within the same top-level limitation (e.g., changing the input for the Text2SQL solution can increase the number of prediction issues) or across top-level limitations (e.g., omitting a metric relaxation can increase the impact of missing labels). Furthermore, some mitigation approaches can have an impact on aspects outside of the taxonomy's scope. For instance, using constraint decoding can increase the latency of the system beyond a pre-defined requirement threshold. Exploring the trade-offs in this large design space and understanding the impact of specific mitigation strategies across and beyond the categories defined in the taxonomy represents an open and challenging problem for future research.

\paragraph{Automatic Categorizations}

To understand the impact of such mitigation strategies in the vast design space of trade-offs, we require automatic categorization methods to apply the proposed taxonomy at different levels of granularity. To go beyond heuristic or ad-hoc methods mentioned throughout the main part of this paper, we need to define benchmarks with datasets and novel evaluation methodologies across all parts of the taxonomy.

\paragraph{Takeaways}

In conclusion, we show that many fundamental challenges in building and evaluating Text2SQL solutions remain significant, particularly when handling ambiguous and unanswerable queries -- issues that even high-quality datasets like Spider sometimes face. A recommendation-based approach, which encourages interactive user guidance and feedback, may prove more suitable for mitigating ambiguity and accommodating real-time clarification. For Text2SQL solutions to be effectively integrated into larger, potentially autonomous systems, they must either be restricted to strictly answerable, unambiguous inputs, where each query maps to a single, valid SQL interpretation, or they must operate within a multi-turn framework, allowing for dynamic user interaction to clarify ambiguity when detected. Finally, fully addressing the limitations across the proposed taxonomy requires new comprehensive benchmarks that assess solutions on unanswerable and multi-answer cases, in addition to new evaluation methodologies tailored to these specific challenges. Such benchmarks would enable the creation of robust algorithms to quantify the amount, and detect which samples are affected by which limitation for a specific Text2SQL solution, evaluation dataset, and evaluation process.



\bibliographystyle{ACM-Reference-Format}
\bibliography{references}


\begin{thebibliography}{40}


\ifx \showCODEN    \undefined \def \showCODEN     #1{\unskip}     \fi
\ifx \showDOI      \undefined \def \showDOI       #1{#1}\fi
\ifx \showISBNx    \undefined \def \showISBNx     #1{\unskip}     \fi
\ifx \showISBNxiii \undefined \def \showISBNxiii  #1{\unskip}     \fi
\ifx \showISSN     \undefined \def \showISSN      #1{\unskip}     \fi
\ifx \showLCCN     \undefined \def \showLCCN      #1{\unskip}     \fi
\ifx \shownote     \undefined \def \shownote      #1{#1}          \fi
\ifx \showarticletitle \undefined \def \showarticletitle #1{#1}   \fi
\ifx \showURL      \undefined \def \showURL       {\relax}        \fi
\providecommand\bibfield[2]{#2}
\providecommand\bibinfo[2]{#2}
\providecommand\natexlab[1]{#1}
\providecommand\showeprint[2][]{arXiv:#2}

\bibitem[\protect\citeauthoryear{Abbe, Bengio, Lotfi, Sandon, and Saremi}{Abbe et~al\mbox{.}}{2024}]%
        {abbe2024far}
\bibfield{author}{\bibinfo{person}{Emmanuel Abbe}, \bibinfo{person}{Samy Bengio}, \bibinfo{person}{Aryo Lotfi}, \bibinfo{person}{Colin Sandon}, {and} \bibinfo{person}{Omid Saremi}.} \bibinfo{year}{2024}\natexlab{}.
\newblock \showarticletitle{How Far Can Transformers Reason? The Locality Barrier and Inductive Scratchpad}.
\newblock \bibinfo{journal}{\emph{arXiv e-prints}} (\bibinfo{year}{2024}), \bibinfo{pages}{arXiv--2406}.
\newblock


\bibitem[\protect\citeauthoryear{Abiteboul, Hull, and Vianu}{Abiteboul et~al\mbox{.}}{1995}]%
        {abiteboul1995foundations}
\bibfield{author}{\bibinfo{person}{Serge Abiteboul}, \bibinfo{person}{Richard Hull}, {and} \bibinfo{person}{Victor Vianu}.} \bibinfo{year}{1995}\natexlab{}.
\newblock \bibinfo{title}{Foundations of Databases: The Logical Level}.
\newblock
\newblock


\bibitem[\protect\citeauthoryear{Ascoli, Kandikonda, and Choi}{Ascoli et~al\mbox{.}}{2024}]%
        {ascoli2024esm+}
\bibfield{author}{\bibinfo{person}{Benjamin Ascoli}, \bibinfo{person}{Ram Kandikonda}, {and} \bibinfo{person}{Jinho~D Choi}.} \bibinfo{year}{2024}\natexlab{}.
\newblock \showarticletitle{ESM+: Modern Insights into Perspective on Text-to-SQL Evaluation in the Age of Large Language Models}.
\newblock \bibinfo{journal}{\emph{arXiv preprint arXiv:2407.07313}} (\bibinfo{year}{2024}).
\newblock


\bibitem[\protect\citeauthoryear{Balachandran, Chen, Joshi, Nushi, Palangi, Salinas, Vineet, Woffinden-Luey, and Yousefi}{Balachandran et~al\mbox{.}}{2024}]%
        {balachandran2024eureka}
\bibfield{author}{\bibinfo{person}{Vidhisha Balachandran}, \bibinfo{person}{Jingya Chen}, \bibinfo{person}{Neel Joshi}, \bibinfo{person}{Besmira Nushi}, \bibinfo{person}{Hamid Palangi}, \bibinfo{person}{Eduardo Salinas}, \bibinfo{person}{Vibhav Vineet}, \bibinfo{person}{James Woffinden-Luey}, {and} \bibinfo{person}{Safoora Yousefi}.} \bibinfo{year}{2024}\natexlab{}.
\newblock \showarticletitle{Eureka: Evaluating and understanding large foundation models}.
\newblock \bibinfo{journal}{\emph{arXiv preprint arXiv:2409.10566}} (\bibinfo{year}{2024}).
\newblock


\bibitem[\protect\citeauthoryear{Bayat, Qian, Han, Sang, Belyi, Khorshidi, Wu, Ilyas, and Li}{Bayat et~al\mbox{.}}{2023}]%
        {bayat2023fleek}
\bibfield{author}{\bibinfo{person}{Farima~Fatahi Bayat}, \bibinfo{person}{Kun Qian}, \bibinfo{person}{Benjamin Han}, \bibinfo{person}{Yisi Sang}, \bibinfo{person}{Anton Belyi}, \bibinfo{person}{Samira Khorshidi}, \bibinfo{person}{Fei Wu}, \bibinfo{person}{Ihab~F Ilyas}, {and} \bibinfo{person}{Yunyao Li}.} \bibinfo{year}{2023}\natexlab{}.
\newblock \showarticletitle{Fleek: Factual error detection and correction with evidence retrieved from external knowledge}.
\newblock \bibinfo{journal}{\emph{arXiv preprint arXiv:2310.17119}} (\bibinfo{year}{2023}).
\newblock


\bibitem[\protect\citeauthoryear{Bhaskar, Tomar, Sathe, and Sarawagi}{Bhaskar et~al\mbox{.}}{2023}]%
        {bhaskar2023benchmarking}
\bibfield{author}{\bibinfo{person}{Adithya Bhaskar}, \bibinfo{person}{Tushar Tomar}, \bibinfo{person}{Ashutosh Sathe}, {and} \bibinfo{person}{Sunita Sarawagi}.} \bibinfo{year}{2023}\natexlab{}.
\newblock \showarticletitle{Benchmarking and Improving Text-to-SQL Generation under Ambiguity}. In \bibinfo{booktitle}{\emph{Proceedings of the 2023 Conference on Empirical Methods in Natural Language Processing}}. \bibinfo{pages}{7053--7074}.
\newblock


\bibitem[\protect\citeauthoryear{Biswal, Patel, Jha, Kamsetty, Liu, Gonzalez, Guestrin, and Zaharia}{Biswal et~al\mbox{.}}{2024}]%
        {biswal2024text2sql}
\bibfield{author}{\bibinfo{person}{Asim Biswal}, \bibinfo{person}{Liana Patel}, \bibinfo{person}{Siddarth Jha}, \bibinfo{person}{Amog Kamsetty}, \bibinfo{person}{Shu Liu}, \bibinfo{person}{Joseph~E Gonzalez}, \bibinfo{person}{Carlos Guestrin}, {and} \bibinfo{person}{Matei Zaharia}.} \bibinfo{year}{2024}\natexlab{}.
\newblock \showarticletitle{Text2SQL is Not Enough: Unifying AI and Databases with TAG}.
\newblock \bibinfo{journal}{\emph{arXiv preprint arXiv:2408.14717}} (\bibinfo{year}{2024}).
\newblock


\bibitem[\protect\citeauthoryear{Chang, Wang, Dong, Pan, Zhu, Li, Lan, Zhang, Jiang, Lilien, Ash, Wang, Wang, Castelli, Ng, and Xiang}{Chang et~al\mbox{.}}{2023}]%
        {chang2023drspider}
\bibfield{author}{\bibinfo{person}{Shuaichen Chang}, \bibinfo{person}{Jun Wang}, \bibinfo{person}{Mingwen Dong}, \bibinfo{person}{Lin Pan}, \bibinfo{person}{Henghui Zhu}, \bibinfo{person}{Alexander~Hanbo Li}, \bibinfo{person}{Wuwei Lan}, \bibinfo{person}{Sheng Zhang}, \bibinfo{person}{Jiarong Jiang}, \bibinfo{person}{Joseph Lilien}, \bibinfo{person}{Steve Ash}, \bibinfo{person}{William~Yang Wang}, \bibinfo{person}{Zhiguo Wang}, \bibinfo{person}{Vittorio Castelli}, \bibinfo{person}{Patrick Ng}, {and} \bibinfo{person}{Bing Xiang}.} \bibinfo{year}{2023}\natexlab{}.
\newblock \showarticletitle{Dr.Spider: A Diagnostic Evaluation Benchmark towards Text-to-{SQL} Robustness}. In \bibinfo{booktitle}{\emph{The Eleventh International Conference on Learning Representations}}.
\newblock
\urldef\tempurl%
\url{https://openreview.net/forum?id=Wc5bmZZU9cy}
\showURL{%
\tempurl}


\bibitem[\protect\citeauthoryear{Chen, Zhang, and Roth}{Chen et~al\mbox{.}}{2024}]%
        {chen2024table}
\bibfield{author}{\bibinfo{person}{Peter~Baile Chen}, \bibinfo{person}{Yi Zhang}, {and} \bibinfo{person}{Dan Roth}.} \bibinfo{year}{2024}\natexlab{}.
\newblock \showarticletitle{Is table retrieval a solved problem? exploring join-aware multi-table retrieval}. In \bibinfo{booktitle}{\emph{Proceedings of the 62nd Annual Meeting of the Association for Computational Linguistics (Volume 1: Long Papers)}}. \bibinfo{pages}{2687--2699}.
\newblock


\bibitem[\protect\citeauthoryear{Chu, Murphy, Roesch, Cheung, and Suciu}{Chu et~al\mbox{.}}{2018}]%
        {chu2018axiomatic}
\bibfield{author}{\bibinfo{person}{Shumo Chu}, \bibinfo{person}{Brendan Murphy}, \bibinfo{person}{Jared Roesch}, \bibinfo{person}{Alvin Cheung}, {and} \bibinfo{person}{Dan Suciu}.} \bibinfo{year}{2018}\natexlab{}.
\newblock \showarticletitle{Axiomatic Foundations and Algorithms for Deciding Semantic Equivalences of SQL Queries}.
\newblock \bibinfo{journal}{\emph{Proceedings of the VLDB Endowment}} \bibinfo{volume}{11}, \bibinfo{number}{11} (\bibinfo{year}{2018}).
\newblock


\bibitem[\protect\citeauthoryear{Dawid and Skene}{Dawid and Skene}{1979}]%
        {dawid1979maximum}
\bibfield{author}{\bibinfo{person}{Alexander~Philip Dawid} {and} \bibinfo{person}{Allan~M Skene}.} \bibinfo{year}{1979}\natexlab{}.
\newblock \showarticletitle{Maximum likelihood estimation of observer error-rates using the EM algorithm}.
\newblock \bibinfo{journal}{\emph{Journal of the Royal Statistical Society: Series C (Applied Statistics)}} \bibinfo{volume}{28}, \bibinfo{number}{1} (\bibinfo{year}{1979}), \bibinfo{pages}{20--28}.
\newblock


\bibitem[\protect\citeauthoryear{Floratou, Psallidas, Zhao, Deep, Hagleither, Tan, Cahoon, Alotaibi, Henkel, Singla, et~al\mbox{.}}{Floratou et~al\mbox{.}}{2024}]%
        {floratou2024nl2sql}
\bibfield{author}{\bibinfo{person}{Avrilia Floratou}, \bibinfo{person}{Fotis Psallidas}, \bibinfo{person}{Fuheng Zhao}, \bibinfo{person}{Shaleen Deep}, \bibinfo{person}{Gunther Hagleither}, \bibinfo{person}{Wangda Tan}, \bibinfo{person}{Joyce Cahoon}, \bibinfo{person}{Rana Alotaibi}, \bibinfo{person}{Jordan Henkel}, \bibinfo{person}{Abhik Singla}, {et~al\mbox{.}}} \bibinfo{year}{2024}\natexlab{}.
\newblock \showarticletitle{Nl2sql is a solved problem... not!}. In \bibinfo{booktitle}{\emph{CIDR}}.
\newblock


\bibitem[\protect\citeauthoryear{Geng, Josifoski, Peyrard, and West}{Geng et~al\mbox{.}}{2023}]%
        {geng2023grammar}
\bibfield{author}{\bibinfo{person}{Saibo Geng}, \bibinfo{person}{Martin Josifoski}, \bibinfo{person}{Maxime Peyrard}, {and} \bibinfo{person}{Robert West}.} \bibinfo{year}{2023}\natexlab{}.
\newblock \showarticletitle{Grammar-Constrained Decoding for Structured NLP Tasks without Finetuning}. In \bibinfo{booktitle}{\emph{The 2023 Conference on Empirical Methods in Natural Language Processing}}.
\newblock


\bibitem[\protect\citeauthoryear{Gkini, Belmpas, Koutrika, and Ioannidis}{Gkini et~al\mbox{.}}{2021}]%
        {gkini2021depth}
\bibfield{author}{\bibinfo{person}{Orest Gkini}, \bibinfo{person}{Theofilos Belmpas}, \bibinfo{person}{Georgia Koutrika}, {and} \bibinfo{person}{Yannis Ioannidis}.} \bibinfo{year}{2021}\natexlab{}.
\newblock \showarticletitle{An in-depth benchmarking of text-to-sql systems}. In \bibinfo{booktitle}{\emph{Proceedings of the 2021 International Conference on Management of Data}}. \bibinfo{pages}{632--644}.
\newblock


\bibitem[\protect\citeauthoryear{Goel, Gu, Li, and R{\'e}}{Goel et~al\mbox{.}}{2020}]%
        {goel2020model}
\bibfield{author}{\bibinfo{person}{Karan Goel}, \bibinfo{person}{Albert Gu}, \bibinfo{person}{Yixuan Li}, {and} \bibinfo{person}{Christopher R{\'e}}.} \bibinfo{year}{2020}\natexlab{}.
\newblock \showarticletitle{Model patching: Closing the subgroup performance gap with data augmentation}.
\newblock \bibinfo{journal}{\emph{arXiv preprint arXiv:2008.06775}} (\bibinfo{year}{2020}).
\newblock


\bibitem[\protect\citeauthoryear{Ilyas and Rekatsinas}{Ilyas and Rekatsinas}{2022}]%
        {ilyas2022machine}
\bibfield{author}{\bibinfo{person}{Ihab~F Ilyas} {and} \bibinfo{person}{Theodoros Rekatsinas}.} \bibinfo{year}{2022}\natexlab{}.
\newblock \showarticletitle{Machine learning and data cleaning: Which serves the other?}
\newblock \bibinfo{journal}{\emph{ACM Journal of Data and Information Quality (JDIQ)}} \bibinfo{volume}{14}, \bibinfo{number}{3} (\bibinfo{year}{2022}), \bibinfo{pages}{1--11}.
\newblock


\bibitem[\protect\citeauthoryear{Jusoh}{Jusoh}{2018}]%
        {jusoh2018study}
\bibfield{author}{\bibinfo{person}{Shaidah Jusoh}.} \bibinfo{year}{2018}\natexlab{}.
\newblock \showarticletitle{A study on NLP applications and ambiguity problems.}
\newblock \bibinfo{journal}{\emph{Journal of Theoretical \& Applied Information Technology}} \bibinfo{volume}{96}, \bibinfo{number}{6} (\bibinfo{year}{2018}).
\newblock


\bibitem[\protect\citeauthoryear{Koutrika}{Koutrika}{2024}]%
        {koutrika2024natural}
\bibfield{author}{\bibinfo{person}{Georgia Koutrika}.} \bibinfo{year}{2024}\natexlab{}.
\newblock \showarticletitle{Natural Language Data Interfaces: A Data Access Odyssey (Invited Talk)}. In \bibinfo{booktitle}{\emph{27th International Conference on Database Theory (ICDT 2024)}}. Schloss Dagstuhl--Leibniz-Zentrum f{\"u}r Informatik.
\newblock


\bibitem[\protect\citeauthoryear{Li, Zhang, Liu, Fan, Zhang, Zhu, Wei, Pan, Li, and Chen}{Li et~al\mbox{.}}{2024b}]%
        {li2024codes}
\bibfield{author}{\bibinfo{person}{Haoyang Li}, \bibinfo{person}{Jing Zhang}, \bibinfo{person}{Hanbing Liu}, \bibinfo{person}{Ju Fan}, \bibinfo{person}{Xiaokang Zhang}, \bibinfo{person}{Jun Zhu}, \bibinfo{person}{Renjie Wei}, \bibinfo{person}{Hongyan Pan}, \bibinfo{person}{Cuiping Li}, {and} \bibinfo{person}{Hong Chen}.} \bibinfo{year}{2024}\natexlab{b}.
\newblock \showarticletitle{Codes: Towards building open-source language models for text-to-sql}.
\newblock \bibinfo{journal}{\emph{Proceedings of the ACM on Management of Data}} \bibinfo{volume}{2}, \bibinfo{number}{3} (\bibinfo{year}{2024}), \bibinfo{pages}{1--28}.
\newblock


\bibitem[\protect\citeauthoryear{Li, Hui, Qu, Yang, Li, Li, Wang, Qin, Geng, Huo, et~al\mbox{.}}{Li et~al\mbox{.}}{2024a}]%
        {li2024can}
\bibfield{author}{\bibinfo{person}{Jinyang Li}, \bibinfo{person}{Binyuan Hui}, \bibinfo{person}{Ge Qu}, \bibinfo{person}{Jiaxi Yang}, \bibinfo{person}{Binhua Li}, \bibinfo{person}{Bowen Li}, \bibinfo{person}{Bailin Wang}, \bibinfo{person}{Bowen Qin}, \bibinfo{person}{Ruiying Geng}, \bibinfo{person}{Nan Huo}, {et~al\mbox{.}}} \bibinfo{year}{2024}\natexlab{a}.
\newblock \showarticletitle{Can llm already serve as a database interface? a big bench for large-scale database grounded text-to-sqls}.
\newblock \bibinfo{journal}{\emph{Advances in Neural Information Processing Systems}}  \bibinfo{volume}{36} (\bibinfo{year}{2024}).
\newblock


\bibitem[\protect\citeauthoryear{Mitsopoulou and Koutrika}{Mitsopoulou and Koutrika}{2025}]%
        {mitsopoulou2024analysis}
\bibfield{author}{\bibinfo{person}{Anna Mitsopoulou} {and} \bibinfo{person}{Georgia Koutrika}.} \bibinfo{year}{2025}\natexlab{}.
\newblock \showarticletitle{Analysis of Text-to-SQL Benchmarks: Limitations, Challenges and Opportunities}.
\newblock  (\bibinfo{year}{2025}).
\newblock


\bibitem[\protect\citeauthoryear{Northcutt, Athalye, and Mueller}{Northcutt et~al\mbox{.}}{2021}]%
        {northcutt2021pervasive}
\bibfield{author}{\bibinfo{person}{Curtis~G Northcutt}, \bibinfo{person}{Anish Athalye}, {and} \bibinfo{person}{Jonas Mueller}.} \bibinfo{year}{2021}\natexlab{}.
\newblock \showarticletitle{Pervasive label errors in test sets destabilize machine learning benchmarks}.
\newblock \bibinfo{journal}{\emph{arXiv preprint arXiv:2103.14749}} (\bibinfo{year}{2021}).
\newblock


\bibitem[\protect\citeauthoryear{Patel, Jha, Guestrin, and Zaharia}{Patel et~al\mbox{.}}{2024}]%
        {patel2024lotus}
\bibfield{author}{\bibinfo{person}{Liana Patel}, \bibinfo{person}{Siddharth Jha}, \bibinfo{person}{Carlos Guestrin}, {and} \bibinfo{person}{Matei Zaharia}.} \bibinfo{year}{2024}\natexlab{}.
\newblock \showarticletitle{Lotus: Enabling semantic queries with llms over tables of unstructured and structured data}.
\newblock \bibinfo{journal}{\emph{arXiv preprint arXiv:2407.11418}} (\bibinfo{year}{2024}).
\newblock


\bibitem[\protect\citeauthoryear{Radford, Wu, Child, Luan, Amodei, Sutskever, et~al\mbox{.}}{Radford et~al\mbox{.}}{2019}]%
        {radford2019language}
\bibfield{author}{\bibinfo{person}{Alec Radford}, \bibinfo{person}{Jeffrey Wu}, \bibinfo{person}{Rewon Child}, \bibinfo{person}{David Luan}, \bibinfo{person}{Dario Amodei}, \bibinfo{person}{Ilya Sutskever}, {et~al\mbox{.}}} \bibinfo{year}{2019}\natexlab{}.
\newblock \showarticletitle{Language models are unsupervised multitask learners}.
\newblock \bibinfo{journal}{\emph{OpenAI blog}} \bibinfo{volume}{1}, \bibinfo{number}{8} (\bibinfo{year}{2019}), \bibinfo{pages}{9}.
\newblock


\bibitem[\protect\citeauthoryear{Rajkumar, Li, and Bahdanau}{Rajkumar et~al\mbox{.}}{2022}]%
        {rajkumar2022evaluating}
\bibfield{author}{\bibinfo{person}{Nitarshan Rajkumar}, \bibinfo{person}{Raymond Li}, {and} \bibinfo{person}{Dzmitry Bahdanau}.} \bibinfo{year}{2022}\natexlab{}.
\newblock \showarticletitle{Evaluating the text-to-sql capabilities of large language models}.
\newblock \bibinfo{journal}{\emph{arXiv preprint arXiv:2204.00498}} (\bibinfo{year}{2022}).
\newblock


\bibitem[\protect\citeauthoryear{Ratner, Bach, Ehrenberg, Fries, Wu, and R{\'e}}{Ratner et~al\mbox{.}}{2020}]%
        {ratner2020snorkel}
\bibfield{author}{\bibinfo{person}{Alexander Ratner}, \bibinfo{person}{Stephen~H Bach}, \bibinfo{person}{Henry Ehrenberg}, \bibinfo{person}{Jason Fries}, \bibinfo{person}{Sen Wu}, {and} \bibinfo{person}{Christopher R{\'e}}.} \bibinfo{year}{2020}\natexlab{}.
\newblock \showarticletitle{Snorkel: rapid training data creation with weak supervision}.
\newblock \bibinfo{journal}{\emph{The VLDB Journal}} \bibinfo{volume}{29}, \bibinfo{number}{2} (\bibinfo{year}{2020}), \bibinfo{pages}{709--730}.
\newblock


\bibitem[\protect\citeauthoryear{Rawte, Sheth, and Das}{Rawte et~al\mbox{.}}{2023}]%
        {rawte2023survey}
\bibfield{author}{\bibinfo{person}{Vipula Rawte}, \bibinfo{person}{Amit Sheth}, {and} \bibinfo{person}{Amitava Das}.} \bibinfo{year}{2023}\natexlab{}.
\newblock \showarticletitle{A survey of hallucination in large foundation models}.
\newblock \bibinfo{journal}{\emph{arXiv preprint arXiv:2309.05922}} (\bibinfo{year}{2023}).
\newblock


\bibitem[\protect\citeauthoryear{Renggli, Rimanic, G{\"u}rel, Karla{\v{s}}, Wu, and Zhang}{Renggli et~al\mbox{.}}{2021a}]%
        {renggli2021data}
\bibfield{author}{\bibinfo{person}{Cedric Renggli}, \bibinfo{person}{Luka Rimanic}, \bibinfo{person}{Nezihe~Merve G{\"u}rel}, \bibinfo{person}{Bojan Karla{\v{s}}}, \bibinfo{person}{Wentao Wu}, {and} \bibinfo{person}{Ce Zhang}.} \bibinfo{year}{2021}\natexlab{a}.
\newblock \showarticletitle{A Data Quality-Driven View of MLOps}.
\newblock \bibinfo{journal}{\emph{IEEE Data Engineering Bulletin}} \bibinfo{volume}{44}, \bibinfo{number}{1} (\bibinfo{year}{2021}), \bibinfo{pages}{11--23}.
\newblock


\bibitem[\protect\citeauthoryear{Renggli, Rimanic, Hollenstein, and Zhang}{Renggli et~al\mbox{.}}{2021b}]%
        {renggli2021evaluating}
\bibfield{author}{\bibinfo{person}{Cedric Renggli}, \bibinfo{person}{Luka Rimanic}, \bibinfo{person}{Nora Hollenstein}, {and} \bibinfo{person}{Ce Zhang}.} \bibinfo{year}{2021}\natexlab{b}.
\newblock \showarticletitle{Evaluating Bayes error estimators on real-world datasets with FeeBee}.
\newblock \bibinfo{journal}{\emph{arXiv preprint arXiv:2108.13034}} (\bibinfo{year}{2021}).
\newblock


\bibitem[\protect\citeauthoryear{Sahoo, Singh, Saha, Jain, Mondal, and Chadha}{Sahoo et~al\mbox{.}}{2024}]%
        {sahoo2024systematic}
\bibfield{author}{\bibinfo{person}{Pranab Sahoo}, \bibinfo{person}{Ayush~Kumar Singh}, \bibinfo{person}{Sriparna Saha}, \bibinfo{person}{Vinija Jain}, \bibinfo{person}{Samrat Mondal}, {and} \bibinfo{person}{Aman Chadha}.} \bibinfo{year}{2024}\natexlab{}.
\newblock \showarticletitle{A systematic survey of prompt engineering in large language models: Techniques and applications}.
\newblock \bibinfo{journal}{\emph{arXiv preprint arXiv:2402.07927}} (\bibinfo{year}{2024}).
\newblock


\bibitem[\protect\citeauthoryear{Shankar, Zamfirescu-Pereira, Hartmann, Parameswaran, and Arawjo}{Shankar et~al\mbox{.}}{2024}]%
        {shankar2024validates}
\bibfield{author}{\bibinfo{person}{Shreya Shankar}, \bibinfo{person}{JD Zamfirescu-Pereira}, \bibinfo{person}{Bj{\"o}rn Hartmann}, \bibinfo{person}{Aditya Parameswaran}, {and} \bibinfo{person}{Ian Arawjo}.} \bibinfo{year}{2024}\natexlab{}.
\newblock \showarticletitle{Who validates the validators? aligning llm-assisted evaluation of llm outputs with human preferences}. In \bibinfo{booktitle}{\emph{Proceedings of the 37th Annual ACM Symposium on User Interface Software and Technology}}. \bibinfo{pages}{1--14}.
\newblock


\bibitem[\protect\citeauthoryear{Sidi, Panahy, Affendey, Jabar, Ibrahim, and Mustapha}{Sidi et~al\mbox{.}}{2012}]%
        {sidi2012data}
\bibfield{author}{\bibinfo{person}{Fatimah Sidi}, \bibinfo{person}{Payam Hassany~Shariat Panahy}, \bibinfo{person}{Lilly~Suriani Affendey}, \bibinfo{person}{Marzanah~A Jabar}, \bibinfo{person}{Hamidah Ibrahim}, {and} \bibinfo{person}{Aida Mustapha}.} \bibinfo{year}{2012}\natexlab{}.
\newblock \showarticletitle{Data quality: A survey of data quality dimensions}. In \bibinfo{booktitle}{\emph{2012 International Conference on Information Retrieval \& Knowledge Management}}. IEEE, \bibinfo{pages}{300--304}.
\newblock


\bibitem[\protect\citeauthoryear{Wang, Gao, Li, and Lou}{Wang et~al\mbox{.}}{2023}]%
        {wang2023know}
\bibfield{author}{\bibinfo{person}{Bing Wang}, \bibinfo{person}{Yan Gao}, \bibinfo{person}{Zhoujun Li}, {and} \bibinfo{person}{Jian-Guang Lou}.} \bibinfo{year}{2023}\natexlab{}.
\newblock \showarticletitle{Know what I don’t know: Handling ambiguous and unknown questions for text-to-sql}. In \bibinfo{booktitle}{\emph{Findings of the Association for Computational Linguistics: ACL 2023}}. \bibinfo{pages}{5701--5714}.
\newblock


\bibitem[\protect\citeauthoryear{Wei, Wang, Schuurmans, Bosma, Xia, Chi, Le, Zhou, et~al\mbox{.}}{Wei et~al\mbox{.}}{2022}]%
        {wei2022chain}
\bibfield{author}{\bibinfo{person}{Jason Wei}, \bibinfo{person}{Xuezhi Wang}, \bibinfo{person}{Dale Schuurmans}, \bibinfo{person}{Maarten Bosma}, \bibinfo{person}{Fei Xia}, \bibinfo{person}{Ed Chi}, \bibinfo{person}{Quoc~V Le}, \bibinfo{person}{Denny Zhou}, {et~al\mbox{.}}} \bibinfo{year}{2022}\natexlab{}.
\newblock \showarticletitle{Chain-of-thought prompting elicits reasoning in large language models}.
\newblock \bibinfo{journal}{\emph{Advances in neural information processing systems}}  \bibinfo{volume}{35} (\bibinfo{year}{2022}), \bibinfo{pages}{24824--24837}.
\newblock


\bibitem[\protect\citeauthoryear{Wu, Zhao, Yasunaga, Huang, Cao, Huang, Ioannidis, Subbian, Zou, and Leskovec}{Wu et~al\mbox{.}}{2024}]%
        {wu2024stark}
\bibfield{author}{\bibinfo{person}{Shirley Wu}, \bibinfo{person}{Shiyu Zhao}, \bibinfo{person}{Michihiro Yasunaga}, \bibinfo{person}{Kexin Huang}, \bibinfo{person}{Kaidi Cao}, \bibinfo{person}{Qian Huang}, \bibinfo{person}{Vassilis~N Ioannidis}, \bibinfo{person}{Karthik Subbian}, \bibinfo{person}{James Zou}, {and} \bibinfo{person}{Jure Leskovec}.} \bibinfo{year}{2024}\natexlab{}.
\newblock \showarticletitle{STaRK: Benchmarking LLM Retrieval on Textual and Relational Knowledge Bases}.
\newblock \bibinfo{journal}{\emph{arXiv preprint arXiv:2404.13207}} (\bibinfo{year}{2024}).
\newblock


\bibitem[\protect\citeauthoryear{Yu, Zhang, Yang, Yasunaga, Wang, Li, Ma, Li, Yao, Roman, et~al\mbox{.}}{Yu et~al\mbox{.}}{2018}]%
        {yu2018spider}
\bibfield{author}{\bibinfo{person}{Tao Yu}, \bibinfo{person}{Rui Zhang}, \bibinfo{person}{Kai Yang}, \bibinfo{person}{Michihiro Yasunaga}, \bibinfo{person}{Dongxu Wang}, \bibinfo{person}{Zifan Li}, \bibinfo{person}{James Ma}, \bibinfo{person}{Irene Li}, \bibinfo{person}{Qingning Yao}, \bibinfo{person}{Shanelle Roman}, {et~al\mbox{.}}} \bibinfo{year}{2018}\natexlab{}.
\newblock \showarticletitle{Spider: A Large-Scale Human-Labeled Dataset for Complex and Cross-Domain Semantic Parsing and Text-to-SQL Task}. In \bibinfo{booktitle}{\emph{Proceedings of the 2018 Conference on Empirical Methods in Natural Language Processing}}. \bibinfo{pages}{3911--3921}.
\newblock


\bibitem[\protect\citeauthoryear{Zhang, Ye, Du, Hu, Li, Yang, Liu, Zhao, Li, and Mao}{Zhang et~al\mbox{.}}{2024}]%
        {zhang2024benchmarking}
\bibfield{author}{\bibinfo{person}{Bin Zhang}, \bibinfo{person}{Yuxiao Ye}, \bibinfo{person}{Guoqing Du}, \bibinfo{person}{Xiaoru Hu}, \bibinfo{person}{Zhishuai Li}, \bibinfo{person}{Sun Yang}, \bibinfo{person}{Chi~Harold Liu}, \bibinfo{person}{Rui Zhao}, \bibinfo{person}{Ziyue Li}, {and} \bibinfo{person}{Hangyu Mao}.} \bibinfo{year}{2024}\natexlab{}.
\newblock \showarticletitle{Benchmarking the text-to-sql capability of large language models: A comprehensive evaluation}.
\newblock \bibinfo{journal}{\emph{arXiv preprint arXiv:2403.02951}} (\bibinfo{year}{2024}).
\newblock


\bibitem[\protect\citeauthoryear{Zhang and Gao}{Zhang and Gao}{2023}]%
        {zhang2023towards}
\bibfield{author}{\bibinfo{person}{Xuan Zhang} {and} \bibinfo{person}{Wei Gao}.} \bibinfo{year}{2023}\natexlab{}.
\newblock \showarticletitle{Towards LLM-based Fact Verification on News Claims with a Hierarchical Step-by-Step Prompting Method}. In \bibinfo{booktitle}{\emph{Proceedings of the 13th International Joint Conference on Natural Language Processing and the 3rd Conference of the Asia-Pacific Chapter of the Association for Computational Linguistics (Volume 1: Long Papers)}}. \bibinfo{pages}{996--1011}.
\newblock


\bibitem[\protect\citeauthoryear{Zhao, Lim, Ahmad, Agrawal, and Abbadi}{Zhao et~al\mbox{.}}{2023}]%
        {zhao2023llm}
\bibfield{author}{\bibinfo{person}{Fuheng Zhao}, \bibinfo{person}{Lawrence Lim}, \bibinfo{person}{Ishtiyaque Ahmad}, \bibinfo{person}{Divyakant Agrawal}, {and} \bibinfo{person}{Amr~El Abbadi}.} \bibinfo{year}{2023}\natexlab{}.
\newblock \showarticletitle{LLM-SQL-Solver: Can LLMs Determine SQL Equivalence?}
\newblock \bibinfo{journal}{\emph{arXiv preprint arXiv:2312.10321}} (\bibinfo{year}{2023}).
\newblock


\bibitem[\protect\citeauthoryear{Zhong, Yu, and Klein}{Zhong et~al\mbox{.}}{2020}]%
        {zhong2020semantic}
\bibfield{author}{\bibinfo{person}{Ruiqi Zhong}, \bibinfo{person}{Tao Yu}, {and} \bibinfo{person}{Dan Klein}.} \bibinfo{year}{2020}\natexlab{}.
\newblock \showarticletitle{Semantic Evaluation for Text-to-SQL with Distilled Test Suites}. In \bibinfo{booktitle}{\emph{Proceedings of the 2020 Conference on Empirical Methods in Natural Language Processing (EMNLP)}}. \bibinfo{pages}{396--411}.
\newblock


\end{thebibliography}

\newpage
\onecolumn
\appendix

\section{Noisy Samples Detector}
\label{app:evluation_data_issues}

To detect potentially noisy samples we query multiple different models available via public APIs. We next provide the prompt used to prompt these models as well as additional examples to the ones given in the main body of this work covering data issues for the popular benchmark Spider.

\subsection{Prompt and result extraction}

We use the following prompt for any LLM, to produce multiple variants for a fixed input consisting of an NL description and a serialized version of a schema:

\begin{tcolorbox}[width=\linewidth, colback=white!95!black]
\begin{prompt}
  You are an expert in writing SQL statements and assessing how complex or ambiguous they are. By ambiguity I mean one of three things:

    1) The natural language statements may contain words that do not make it clear to which table or column the statement refers to
    2) The natural language statement may contain words that do not make it clear to which value the statement refers to 
    3) The natural language statement does clarify if the operation should return a set of unique items or a collection where duplicates are allowed
    
  For the the statement "{NL_DESCRIPTION}" for schema: "{SERIALIZED_SCHEMA}"
  Can you propose up at most three possible SQL statements that capture the ambiguity in the statement if present? You can generate less than three if there is little or no ambiguity.
\end{prompt}
\end{tcolorbox}

Extraction is performed via regex matching and finding corresponding SQL blocks: \verb~`{3}(?:sql|SQL)([\s\S]+?)`{3}~.

\subsection{Aggregation Strategies}

As mentioned in the main body of this work, we use two different aggregation strategies. We fix commonly used match functions with different hyper-parameters. To identify inaccurate labels, we take the union of all provided SQL variants across different LLMs, and check whether the provided GT label is contained in that set. To find other noisy samples (possibly overlapping), we take a majority voting approach, where the variant is contained in the set of alternative SQL variants if at least half of the LLMs (including the LLM) proposed this approach. Note that for this we need at least 3 LLMs to provide alternative variants.

\subsection{Examples - Innacurate Labels}

We identify 43 samples (roughly 2\%) samples in the test split of Spider likely to have inaccurate labels.

\begin{longtable}{p{0.15\linewidth}p{0.22\linewidth}p{0.33\linewidth}p{0.2\linewidth}}
\toprule
Database                            & NL Description                                                                                                           & GT SQL Query                                                                                                                                                                                                                                                                                                                                                    & Explanation                                                                                                                                                           \\ \midrule
aan\_1                              & Find the number of authors who did not publish any paper that is cited more than 50 times.                               & SELECT count(*) FROM Author WHERE Author\_id NOT IN ( SELECT T2.author\_id FROM Citation AS T1 JOIN Author\_list AS T2 ON T1.cited\_paper\_id  =  T2.paper\_id GROUP BY T1.cited\_paper\_id HAVING count(DISTINCT T1.paper\_id)  \textgreater  50)                                                                                                              & Group and count over full joins wrong (accounts for multiple authors per paper), hence author with id `494' is counted in and should not because of paper `N09-1003' \\
aan\_1                              & Find the titles and paper IDs for papers which have Mckeown, Kathleen or Rambow, Owen in author list.                    & SELECT DISTINCT T1.title ,  T1.paper\_id FROM Paper AS T1 JOIN Author\_list AS T2 ON T1.paper\_id  =  T2.paper\_id JOIN Author AS T3 ON T2.author\_id  =  T3.author\_id WHERE  T3.name LIKE "\%Mckeown ,  Kathleen\%" OR T3.name LIKE "\%Rambow ,  Owen\%"                                                                                                      & Value comparison                                                                                                                                                      \\
aan\_1                              & What are the titles and paper ids for papers that have Mckeown, Kathleen or Rambow, Owen in their author list?           & SELECT DISTINCT T1.title ,  T1.paper\_id FROM Paper AS T1 JOIN Author\_list AS T2 ON T1.paper\_id  =  T2.paper\_id JOIN Author AS T3 ON T2.author\_id  =  T3.author\_id WHERE  T3.name LIKE "\%Mckeown ,  Kathleen\%" OR T3.name LIKE "\%Rambow ,  Owen\%"                                                                                                      & Value comparison                                                                                                                                                      \\
aan\_1                              & Count the number of papers Columbia University had during or prior to 2009 .                                             & select count(distinct t1.paper\_id) from paper as t1 join author\_list as t2 on t1.paper\_id  =  t2.paper\_id join affiliation as t3 on t2.affiliation\_id  =  t3.affiliation\_id where t1.year  \textless{}=  2009 and t3.name = "columbia university"                                                                                                         & Value comparison                                                                                                                                                      \\
aan\_1                              & How many papers does Columbia University have in or before 2009 ?                                                        & select count(distinct t1.paper\_id) from paper as t1 join author\_list as t2 on t1.paper\_id  =  t2.paper\_id join affiliation as t3 on t2.affiliation\_id  =  t3.affiliation\_id where t1.year  \textless{}=  2009 and t3.name = "columbia university"                                                                                                         & Value comparison                                                                                                                                                      \\
aan\_1                              & What are the venues and years where Columbia University had papers ?                                                     & select distinct t1.venue ,  t1.year from paper as t1 join author\_list as t2 on t1.paper\_id  =  t2.paper\_id join affiliation as t3 on t2.affiliation\_id  =  t3.affiliation\_id where t3.name = "columbia university"                                                                                                                                         & Value comparison                                                                                                                                                      \\
aan\_1                              & Which venues and years did Columbia University have papers ?                                                             & select distinct t1.venue ,  t1.year from paper as t1 join author\_list as t2 on t1.paper\_id  =  t2.paper\_id join affiliation as t3 on t2.affiliation\_id  =  t3.affiliation\_id where t3.name = "columbia university"                                                                                                                                         & Value comparison                                                                                                                                                      \\
bakery\_1                           & What is the most expensive cake and its flavor?                                                                          & SELECT id ,  flavor FROM goods WHERE food  =  "Cake" ORDER BY price DESC LIMIT 1                                                                                                                                                                                                                                                                                & More than one min / Which Columns                                                                                                                                     \\
bbc\_channels                       & Find the id and name of the channel that is not directed by Hank Baskett.                                                & SELECT t1.name ,  t1.channel\_id FROM channel AS t1 JOIN director\_admin AS t2 ON t1.channel\_id  =  t2.channel\_id JOIN director AS t3 ON t2.director\_id  =  t3.director\_id WHERE t3.name != "Hank Baskett"                                                                                                                                                  & Channels without directors?                                                                                                                                           \\
book\_1                             & What are the titles of books that have a sale price equal to the lowest sale price across all books ?                    & select title from book order by saleprice asc limit 1                                                                                                                                                                                                                                                                                                           & Wrong presupposition (Top-K)                                                                                                                                                       \\
book\_press                         & What are the 5 best books in terms of sale amount? Give me their titles and release dates.                               & SELECT title ,  release\_date FROM book ORDER BY sale\_amount DESC LIMIT 5                                                                                                                                                                                                                                                                                      & Sale\_Amount need to be casted                                                                                                                                        \\
car\_racing                         & Which make has more than one team?                                                                                       & SELECT make FROM team GROUP BY team HAVING count(*)  \textgreater  1                                                                                                                                                                                                                                                                                            & Wrong group by attribute                                                                                                                                              \\
car\_racing                         & Find the make that has more than one team.                                                                               & SELECT make FROM team GROUP BY team HAVING count(*)  \textgreater  1                                                                                                                                                                                                                                                                                            & Wrong group by attribute                                                                                                                                              \\
club\_leader                        & List the names of members whose nationality is not ``England'' .                                                       & select name from member where nationality != "england"                                                                                                                                                                                                                                                                                                          & String comparison wrong                                                                                                                                               \\
conference                          & What is the maximum and minimum age of all staff from the United States?                                                 & SELECT max(age) ,  min(age) FROM staff                                                                                                                                                                                                                                                                                                                          & Missing filter condition                                                                                                                                              \\
\begin{tabular}{@{}l@{}}cre\_Doc\_and\_\\collections\end{tabular}          & List all owner of documents that is related to documents owned by Braeden.                                               & SELECT DISTINCT OWNER FROM Document\_Subset\_Members AS T1 JOIN Document\_Objects AS T2 ON T1.Related\_Document\_Object\_ID  =  T2.Document\_Object\_ID WHERE T2.Owner  =  `Braeden';                                                                                                                                                                           & Missing new join to document\_object based on related ids                                                                                                             \\
\begin{tabular}{@{}l@{}}cre\_Doc\_and\_\\collections\end{tabular}          & What are the document object ids of the related to the document owned by Ransom ?                                        & select t1.document\_object\_id from document\_subset\_members as t1 join document\_objects as t2 on t1.document\_object\_id  =  t2.document\_object\_id where t2.owner  =  `ransom'                                                                                                                                                                             & Value comparison                                                                                                                                                      \\
\begin{tabular}{@{}l@{}}cre\_Doc\_and\_\\collections\end{tabular}          & What are the document object ids of the objects owned by Marlin?                                                         & SELECT Parent\_Document\_Object\_ID FROM Document\_Objects WHERE OWNER  =  `Marlin'                                                                                                                                                                                                                                                                             & wrong projection                                                                                                                                                      \\
\begin{tabular}{@{}l@{}}cre\_Doc\_and\_\\collections\end{tabular}          & What is the number of document object ids in the collection named Best?                                                  & SELECT T2.Document\_Object\_ID FROM Collections AS T1 JOIN Documents\_in\_Collections AS T2 ON T1.Collection\_ID = T2.Collection\_ID WHERE T1.Collection\_Name = "Best";                                                                                                                                                                                        & Missing count                                                                                                                                                         \\
\begin{tabular}{@{}l@{}}cre\_Doc\_and\_\\collections\end{tabular}          & List the Collection Name that document owned by `Ransom' belong to .                                                    & select t1.collection\_name from collections as t1 join documents\_in\_collections as t2 on t1.collection\_id = t2.collection\_id join document\_objects as t3 on t2.document\_object\_id  =  t3.document\_object\_id where t3.owner  =  `ransom'                                                                                                                & Value comparison                                                                                                                                                      \\
\begin{tabular}{@{}l@{}}cre\_Students\_Infor-\\mation\_Systems\end{tabular} & When was the transcript issued for the student with loan of maximum value?                                               & SELECT T1.date\_of\_transcript FROM Transcripts AS T1 JOIN Student\_Loans AS T2 ON T1.student\_id  =  T2.student\_id ORDER BY T2.amount\_of\_loan DESC LIMIT 1                                                                                                                                                                                                  & More than one max                                                                                                                                                     \\
\begin{tabular}{@{}l@{}}cre\_Students\_Infor-\\mation\_Systems\end{tabular} & What is the transcript issuance date for the student with the largest amount of loan?                                    & SELECT T1.date\_of\_transcript FROM Transcripts AS T1 JOIN Student\_Loans AS T2 ON T1.student\_id  =  T2.student\_id ORDER BY T2.amount\_of\_loan DESC LIMIT 1                                                                                                                                                                                                  & More than one max                                                                                                                                                     \\
government\_shift                   & Which customers have integration platform details "Fail" in interactions? Give me the customer details.                  & SELECT DISTINCT t1.customer\_details FROM customers AS t1 JOIN customer\_interactions AS t2 ON t1.customer\_id  =  t2.customer\_id JOIN integration\_platform AS t3 WHERE t3.integration\_platform\_details  =  "Fail"                                                                                                                                          & Missing join condition on T3                                                                                                                                          \\
institution\_sports                 & What are the total enrollment of institutions in city ``Vancouver'' or ``Calgary'' ?                                 & select sum(enrollment) from institution where city  =  "vancouver" or city  =  "calgary"                                                                                                                                                                                                                                                                        & Value comparison                                                                                                                                                      \\
institution\_sports                 & Return all the enrollments of institutions in either the city of Vancouver or the city of Calgary .                      & select sum(enrollment) from institution where city  =  "vancouver" or city  =  "calgary"                                                                                                                                                                                                                                                                        & Value comparison and wrong template                                                                                                                                   \\
movie\_2                            & Show all information of all unrated movies.                                                                              & SELECT * FROM movies WHERE rating  =  `null'                                                                                                                                                                                                                                                                                                                    & Null value comparison                                                                                                                                                 \\
movie\_2                            & What are the names of all cinemas not showing any movies?                                                                & SELECT DISTINCT name FROM MovieTheaters WHERE Movie  =  `null'                                                                                                                                                                                                                                                                                                  & Null value comparison                                                                                                                                                 \\
movie\_2                            & What are the names of movies that do not have any ratings?                                                               & SELECT title FROM movies WHERE rating  =  `null'                                                                                                                                                                                                                                                                                                                & Null value comparison                                                                                                                                                 \\
movie\_2                            & What are the names of the movie theaters that are playing `G' or `PG' rated movies?                                      & SELECT title FROM movies WHERE rating  =  `G' OR rating  =  `PG'                                                                                                                                                                                                                                                                                                & Giving back movie titles instead of cinema titles                                                                                                                     \\
movie\_2                            & Find the titles of movies that don’t have any rating.                                                                    & SELECT title FROM movies WHERE rating  =  `null'                                                                                                                                                                                                                                                                                                                & Null value comparison                                                                                                                                                 \\
movie\_2                            & Find the name of the cinemas that are playing movies with either rating ‘G’ or rating ‘PG’.                              & SELECT title FROM movies WHERE rating  =  `G' OR rating  =  `PG'                                                                                                                                                                                                                                                                                                & Giving back movie titles instead of cinema titles                                                                                                                     \\
planet\_1                           & List all package sent or received by Leo Wong.                                                                           & SELECT DISTINCT T1.PackageNumber FROM PACKAGE AS T1 JOIN Client AS T2 ON T1.Sender  =  T2.AccountNumber OR T1.Recipient  =  T2.AccountNumber WHERE T2.Name = "Leo Wong"                                                                                                                                                                                         & Schema confusing (package number is not an id)                                                                                                                        \\
planet\_1                           & What are the ids for all shipments on the planet Mars that Turanga Leela manages?                                        & SELECT T1.ShipmentID FROM Shipment AS T1 JOIN Planet AS T2 ON T1.Planet = T2.PlanetID JOIN Employee AS T3 ON T3.EmployeeID = T1.Manager WHERE T2.Name = "Mars" OR T3.Name = "Turanga Leela";                                                                                                                                                                    & OR instead of AND                                                                                                                                                     \\
planet\_1                           & What is the package number and weight of the heaviest package that was sent by a client named John or something similar? & SELECT T1.PackageNumber ,  max(T1.Weight) FROM PACKAGE AS T1 JOIN Client AS T2 ON T1.Sender  =  T2.AccountNumber WHERE T2.Name LIKE "John";                                                                                                                                                                                                                     & Null value comparison                                                                                                                                                 \\
planet\_1                           & List all package numbers received by Leo Wong ?                                                                          & select t1.packagenumber from package as t1 join client as t2 on t1.recipient  =  t2.accountnumber where t2.name = "leo wong";                                                                                                                                                                                                                                   & Null value comparison                                                                                                                                                 \\
planet\_1                           & What is the heaviest package sent by the clients which `John' is part of their name? List package number and weight.     & SELECT T1.PackageNumber ,  max(T1.Weight) FROM PACKAGE AS T1 JOIN Client AS T2 ON T1.Sender  =  T2.AccountNumber WHERE T2.Name LIKE "John";                                                                                                                                                                                                                     & Null value comparison                                                                                                                                                 \\
real\_estate\_rentals               & How many properties are there with at least 2 features?                                                                  & SELECT count(*) FROM Properties GROUP BY property\_id HAVING count(*)  \textgreater{}= 2;                                                                                                                                                                                                                                                                       & Missing join and wrong group by attribute                                                                                                                             \\
real\_estate\_rentals               & Count the number of properties with at least two features.                                                               & SELECT count(*) FROM Properties GROUP BY property\_id HAVING count(*)  \textgreater{}= 2;                                                                                                                                                                                                                                                                       & Missing join and wrong group by attribute                                                                                                                             \\
region\_building                    & For each building, return the address of the building and the name of the region it belongs to.                          & SELECT T1.Address ,  T2.Capital FROM building AS T1 JOIN region AS T2 ON T1.Region\_ID  =  T2.Region\_ID                                                                                                                                                                                                                                                        & returns capital name instead of region name                                                                                                                           \\
tv\_shows                           & What is the transmitter of the radio with the largest ERP\_kW?                                                           & SELECT Transmitter FROM radio ORDER BY ERP\_kW DESC LIMIT 1                                                                                                                                                                                                                                                                                                     & Missing cast for property / more than one max                                                                                                                         \\
video\_game                         & What are the titles of games that are played by players from Oklahoma college or Auburn college?                         & SELECT T1.Title FROM game AS T1 JOIN game\_player AS T2 ON T1.Game\_ID  =  T2.Game\_ID JOIN player AS T3 ON T2.Player\_ID  =  T3.Player\_ID WHERE T3.College  =  "Oklahoma" INTERSECT SELECT T1.Title FROM game AS T1 JOIN game\_player AS T2 ON T1.Game\_ID  =  T2.Game\_ID JOIN player AS T3 ON T2.Player\_ID  =  T3.Player\_ID WHERE T3.College  =  "Auburn" & Label noise (AND instead of OR)                                                                                                                                       \\
warehouse\_1                        & What are the different locations of warehouses?                                                                          & SELECT count(DISTINCT LOCATION) FROM warehouses                                                                                                                                                                                                                                                                                                                 & Count instead of projection                                                                                                                                           \\
warehouse\_1                        & Find all distinct locations of warehouses.                                                                               & SELECT count(DISTINCT LOCATION) FROM warehouses                                                                                                                                                                                                                                                                                                                 & Count instead of projection                                                                                                                                           \\ \bottomrule
\end{longtable}

\subsection{Examples - Other Data Issues}

We identify another list of 42 samples, for which the feature and label contain some degree of noise based on manual inspection.

\begin{longtable}{p{0.15\linewidth}p{0.22\linewidth}p{0.33\linewidth}p{0.2\linewidth}}
\toprule
Database                            & NL Description                                                                                                                                        & GT SQL Query                                                                                                                                                                                                                                                                                                                                                       & Explanation                                                             \\ \midrule
aan\_1                              & Find the names of authors who published some paper on NAACL and ACL in the year 2009.                                                                 & SELECT name FROM Author WHERE author\_id IN (SELECT T1.author\_id FROM Author\_list AS T1 JOIN Paper AS T2 ON T1.paper\_id  =  T2.paper\_id WHERE T2.venue  =  "ACL" AND T2.year  =  2009 INTERSECT SELECT T1.author\_id FROM Author\_list AS T1 JOIN Paper AS T2 ON T1.paper\_id  =  T2.paper\_id WHERE T2.venue  =  "NAACL" AND T2.year  =  2009)                & Is `and' in the text a logical AND or a logical OR?                     \\
aan\_1                              & How many reference papers does paper with id D12-1027 have?                                                                                           & SELECT count(*) FROM Citation WHERE paper\_id  =  "D12-1027"                                                                                                                                                                                                                                                                                                       & Is referenced citing or being cited by?                                 \\
address\_1                          & What is the first name and last name of the student living furthest to Linda Smith?                                                                   & SELECT T3.Fname ,  T3.Lname FROM Direct\_distance AS T1 JOIN Student AS T2 ON T1.city1\_code  =  T2.city\_code JOIN Student AS T3 ON T1.city2\_code  =  T3.city\_code WHERE T2.Fname  =  "Linda" AND T2.Lname  =  "Smith" ORDER BY distance DESC LIMIT 1                                                                                                           & What if there is no student in the city the furthest away?              \\
address\_1                          & What is the full name of the student who lives furthest from Linda Smith?                                                                             & SELECT T3.Fname ,  T3.Lname FROM Direct\_distance AS T1 JOIN Student AS T2 ON T1.city1\_code  =  T2.city\_code JOIN Student AS T3 ON T1.city2\_code  =  T3.city\_code WHERE T2.Fname  =  "Linda" AND T2.Lname  =  "Smith" ORDER BY distance DESC LIMIT 1                                                                                                           & Which Columns                                                           \\
address\_1                          & What are the full names of students living in MD?                                                                                                     & SELECT T2.Fname ,  T2.Lname FROM City AS T1 JOIN Student AS T2 ON T1.city\_code  =  T2.city\_code WHERE T1.state  =  "MD"                                                                                                                                                                                                                                          & Which Columns                                                           \\
art\_1                              & What are the full names of artists born after 1850?                                                                                                   & SELECT lname ,  fname FROM artists WHERE birthYear  \textgreater  1850                                                                                                                                                                                                                                                                                             & Which Columns                                                           \\
art\_1                              & What is the name of the oldest painting and where is it located?                                                                                      & SELECT title ,  LOCATION ,  YEAR FROM paintings ORDER BY YEAR LIMIT 1                                                                                                                                                                                                                                                                                              & Which Columns                                                           \\
art\_1                              & In what locations and on what mediums are the paintings created by the artist with the first name Pablo?                                              & SELECT T2.location ,   T2.medium FROM artists AS T1 JOIN paintings AS T2 ON T1.artistID  =  T2.painterID WHERE T1.fname  =  "Pablo"                                                                                                                                                                                                                                & Which Columns                                                           \\
art\_1                              & What are the locations that have works painted before 1885 and after 1930?                                                                            & SELECT DISTINCT LOCATION FROM paintings WHERE YEAR  \textless  1885 INTERSECT SELECT DISTINCT LOCATION FROM paintings WHERE YEAR  \textgreater  1930                                                                                                                                                                                                               & Is `and' in the text a logical AND or a logical OR?                     \\
art\_1                              & Find the locations that have paintings created before 1885 and after 1930?                                                                            & SELECT DISTINCT LOCATION FROM paintings WHERE YEAR  \textless  1885 INTERSECT SELECT DISTINCT LOCATION FROM paintings WHERE YEAR  \textgreater  1930                                                                                                                                                                                                               & Is `and' in the text a logical AND or a logical OR?                     \\
bakery\_1                           & What is the cheapest cookie and its flavor?                                                                                                           & SELECT id ,  flavor FROM goods WHERE food  =  "Cookie" ORDER BY price LIMIT 1                                                                                                                                                                                                                                                                                      & Which Columns                                                           \\
bakery\_1                           & What is the cheapest good with cheese flavor?                                                                                                         & SELECT min(price) FROM goods WHERE flavor  =  "Cheese"                                                                                                                                                                                                                                                                                                             & Which Columns                                                           \\
bakery\_1                           & What are the full names of customers who bought apple flavored Tarts?                                                                                 & SELECT DISTINCT T4.FirstName ,  T4.LastName FROM goods AS T1 JOIN items AS T2 ON T1.id  =  T2.item JOIN receipts AS T3 ON T2.receipt  =  T3.ReceiptNumber JOIN customers AS T4 ON T3.CustomerId  =  T4.id WHERE T1.flavor  =  "Apple" AND T1.food  =  "Tart"                                                                                                       & Which Columns                                                           \\
bakery\_1                           & What is the full name of the customer who visited on the earliest date?                                                                               & SELECT T1.FirstName ,  T1.LastName FROM customers AS T1 JOIN receipts AS T2 ON T1.id  =  T2.CustomerId ORDER BY T2.date LIMIT 1                                                                                                                                                                                                                                    & Which Columns                                                           \\
boat\_1                             & Find boats reserved by Sailor with id 1.                                                                                                              & SELECT DISTINCT bid FROM Reserves WHERE sid = 1                                                                                                                                                                                                                                                                                                                    & Which Columns                                                           \\
book\_press                         & Find the average and maximum yearly profit for each press.                                                                                            & SELECT avg(Year\_Profits\_billion) ,  max(Year\_Profits\_billion) FROM press                                                                                                                                                                                                                                                                                       & Question wrt schema unclear!                                            \\
book\_review                        & What are the maximum and minimum number of chapters for each book?                                                                                    & SELECT max(Chapters) ,  min(Chapters) FROM book                                                                                                                                                                                                                                                                                                                    & Question wrt schema unclear!                                            \\
car\_racing                         & Find the highest and lowest points of drivers.                                                                                                        & SELECT max(Points) ,  min(Points) FROM driver                                                                                                                                                                                                                                                                                                                      & Casting to INT                                                          \\
club\_leader                        & Show the names of club leaders that joined their club before 2018.                                                                                    & SELECT T3.Name ,  T2.Club\_Name FROM club\_leader AS T1 JOIN club AS T2 ON T1.Club\_ID  =  T2.Club\_ID JOIN member AS T3 ON T1.Member\_ID  =  T3.Member\_ID WHERE T1.Year\_Join  \textless  2018                                                                                                                                                                   & Which Columns                                                           \\
\begin{tabular}{@{}l@{}}cre\_Students\_Infor-\\mation\_Systems\end{tabular} & Which students have gone through any event? List the students' biographical data and event date.                                                      & SELECT T1.bio\_data ,  T2.event\_date FROM Students AS T1 JOIN Student\_Events AS T2 ON T1.student\_id  =  T2.student\_id                                                                                                                                                                                                                                          & Which Columns                                                           \\
\begin{tabular}{@{}l@{}}cre\_Students\_Infor-\\mation\_Systems\end{tabular} & Which student has the loan with the minimum value? List the student's biographical information.                                                       & SELECT T1.bio\_data FROM Students AS T1 JOIN Student\_Loans AS T2 ON T1.student\_id  =  T2.student\_id ORDER BY T2.amount\_of\_loan ASC LIMIT 1                                                                                                                                                                                                                    & Which Columns                                                           \\
\begin{tabular}{@{}l@{}}cre\_Students\_Infor-\\mation\_Systems\end{tabular} & What are the biographical data and the date of transcript issuance of each student?                                                                   & SELECT T1.bio\_data ,  T2.date\_of\_transcript FROM Students AS T1 JOIN Transcripts AS T2 ON T1.student\_id  =  T2.student\_id                                                                                                                                                                                                                                     & Which Columns                                                           \\
\begin{tabular}{@{}l@{}}cre\_Students\_Infor-\\mation\_Systems\end{tabular} & Find the details of the teachers who have taught the student with the earliest transcript issuance.                                                   & SELECT T1.teacher\_details FROM Teachers AS T1 JOIN Classes AS T2 ON T1.teacher\_id  =  T2.teacher\_id JOIN Transcripts AS T3 ON T2.student\_id  =  T3.student\_id ORDER BY T3.date\_of\_transcript ASC LIMIT 1                                                                                                                                                    & Which Columns                                                           \\
\begin{tabular}{@{}l@{}}cre\_Students\_Infor-\\mation\_Systems\end{tabular} & What are the address details and biographical information of each student?                                                                            & SELECT T1.address\_details ,  T3.bio\_data FROM Addresses AS T1 JOIN Students\_Addresses AS T2 ON T1.address\_id  =  T2.address\_id JOIN Students AS T3 ON T2.student\_id  =  T3.student\_id                                                                                                                                                                       & Which Columns                                                           \\
\begin{tabular}{@{}l@{}}cre\_Students\_Infor-\\mation\_Systems\end{tabular} & Which students have `Suite' as a substring in their details? Give me their biographical information.                                                  & SELECT bio\_data FROM Students WHERE student\_details LIKE `\%Suite\%'                                                                                                                                                                                                                                                                                             & Which Columns                                                           \\
\begin{tabular}{@{}l@{}}cre\_Students\_Infor-\\mation\_Systems\end{tabular} & List the details for all the pairs of teachers and students who are in the same class.                                                                & SELECT T1.teacher\_details ,  T3.student\_details FROM Teachers AS T1 JOIN Classes AS T2 ON T1.teacher\_id  =  T2.teacher\_id JOIN Students AS T3 ON T2.student\_id  =  T3.student\_id                                                                                                                                                                             & Which Columns                                                           \\
e\_commerce                         & What are the statuses, dates, and shipment dates for all invoices?                                                                                    & SELECT T1.invoice\_status\_code ,  T1.invoice\_date ,  T2.shipment\_date FROM Invoices AS T1 JOIN Shipments AS T2 ON T1.invoice\_number  =  T2.invoice\_number                                                                                                                                                                                                     & What about not shipped shipments (NULL)                                 \\
e\_commerce                         & What is the status code of the items being ordered and shipped and its corresponding shipment tracking number?                                        & SELECT T1.order\_item\_status\_code ,  T3.shipment\_tracking\_number FROM Order\_items AS T1 JOIN Shipment\_Items AS T2 ON T1.order\_item\_id  =  T2.order\_item\_id JOIN Shipments AS T3 ON T2.shipment\_id  =  T3.shipment\_id                                                                                                                                   & Unnormalised data over joins!                                           \\
e\_commerce                         & What is the product name and the color of the ordered items which have been shipped?                                                                  & SELECT T1.product\_name ,  T1.product\_color FROM Products AS T1 JOIN Order\_items AS T2 ON T1.product\_id  =  T2.product\_id JOIN Shipment\_Items AS T3 ON T2.order\_item\_id  =  T3.order\_item\_id JOIN Shipments AS T4 ON T3.shipment\_id  =  T4.shipment\_id                                                                                                  & Unnormalised data over joins!                                           \\
e\_commerce                         & How many products have been shipped?                                                                                                                  & SELECT count(*) FROM Shipment\_Items                                                                                                                                                                                                                                                                                                                               & Unnormalised data over joins!                                           \\
e\_commerce                         & What is the status code of the items have been ordered and shipped, and also what are their shipment tracking numbers?                                & SELECT T1.order\_item\_status\_code ,  T3.shipment\_tracking\_number FROM Order\_items AS T1 JOIN Shipment\_Items AS T2 ON T1.order\_item\_id  =  T2.order\_item\_id JOIN Shipments AS T3 ON T2.shipment\_id  =  T3.shipment\_id                                                                                                                                   & Which Columns (invoice or order status code)                            \\
government\_shift                   & Find details of all the services that have interacted with ``15 ij'' for the the channel details.                                                   & select distinct t1.service\_details from services as t1 join customer\_interactions as t2 on t1.service\_id  =  t2.service\_id join channels as t3 on t2.channel\_id  =  t3.channel\_id where t3.channel\_details  =  "15 ij"                                                                                                                                      & Which Columns                                                           \\
government\_shift                   & Give me the details of all the services that have interacted with the channel with detail "15 ij".                                                    & SELECT DISTINCT t1.service\_details FROM services AS t1 JOIN customer\_interactions AS t2 ON t1.service\_id  =  t2.service\_id JOIN channels AS t3 ON t2.channel\_id  =  t3.channel\_id WHERE t3.channel\_details  =  "15 ij"                                                                                                                                      & Which Columns                                                           \\
government\_shift                   & Count the total number of available customers and services details.                                                                                   & SELECT count(DISTINCT customers\_and\_services\_details) FROM customers\_and\_services                                                                                                                                                                                                                                                                             & Count distinct vs not distinct                                          \\
government\_shift                   & Find the names of all the services which both have been used by customer "Hardy Kutch" and have been rated "bad" in one of the customer interactions. & SELECT DISTINCT t3.service\_details FROM customers AS t1 JOIN customers\_and\_services AS t2 ON t1.customer\_id  =  t2.customer\_id JOIN services AS t3 ON t2.service\_id  =  t3.service\_id JOIN customer\_interactions AS t4 ON t3.service\_id  =  t4.service\_id WHERE t1.customer\_details  =  "Hardy Kutch" AND t4.services\_and\_channels\_details  =  "bad" & Unnormalised and inconsistent data wrt customer and service details     \\
government\_shift                   & How many distinct customer and services details are there?                                                                                            & SELECT count(DISTINCT customers\_and\_services\_details) FROM customers\_and\_services                                                                                                                                                                                                                                                                             & Unnormalised and inconsistent data wrt customer and service details     \\
pilot\_1                            & Count the number of B-52 Bombers owned by pilots under 35.                                                                                            & SELECT count(*) FROM pilotskills WHERE age  \textless  35 AND plane\_name  =  `B-52 Bomber'                                                                                                                                                                                                                                                                        & Unnormalised schema! Are planes and pilots unique by their name or not? \\
pilot\_1                            & Count the number of planes flown by pilots older than 40.                                                                                             & SELECT count(plane\_name) FROM pilotskills WHERE age  \textgreater  40                                                                                                                                                                                                                                                                                             & Unnormalised schema! Are planes and pilots unique by their name or not? \\
pilot\_1                            & How many planes are controlled by the pilots whose age is older than 40?                                                                              & SELECT count(plane\_name) FROM pilotskills WHERE age  \textgreater  40                                                                                                                                                                                                                                                                                             & Unnormalised schema! Are planes and pilots unique by their name or not? \\
real\_estate\_rentals               & Return the full names of users who live in properties that they own.                                                                                  & SELECT first\_name ,  middle\_name ,  last\_name FROM Properties AS T1 JOIN Users AS T2 ON T1.owner\_user\_id  =  T2.user\_id WHERE T1.property\_address\_id  =  T2.user\_address\_id;                                                                                                                                                                             & Which Columns                                                           \\
real\_estate\_rentals               & What is the age range category of the user who made the first search?                                                                                 & SELECT T1.age\_category\_code FROM Users AS T1 JOIN User\_Searches AS T2 ON T1.user\_id  =  T2.user\_id ORDER BY T2.search\_datetime LIMIT 1;                                                                                                                                                                                                                      & Which Columns                                                           \\
restaurant\_bills                   & Find the customer with the highest membership level and return his or her card credit.                                                                & SELECT Card\_Credit FROM customer ORDER BY Level\_of\_Membership DESC LIMIT 1                                                                                                                                                                                                                                                                                      & Which Columns                                                           \\ \bottomrule
\end{longtable}

\end{document}